\documentclass[final,3p,times,twocolumn]{elsarticle}

\usepackage{hyperref}

\journal{Computers and Electronics in Agriculture}

\usepackage{amssymb}
\usepackage{colortbl}
\usepackage{url}
\usepackage{color}
\usepackage[table]{xcolor}
\usepackage{booktabs}
\usepackage{multirow}
\usepackage[caption=false]{subfig}
\usepackage[sort,nocompress]{cite}
\usepackage{stfloats}
\usepackage{float}
\usepackage{textcomp}
\usepackage{xcolor, colortbl}
\usepackage{comment}
\usepackage{pgfplots}
\pgfplotsset{compat=1.17} 

\definecolor{amber}{rgb}{1.0, 0.75, 0.0}
\definecolor{darkgoldenrod}{rgb}{0.72, 0.53, 0.04}

\newcommand{\HM}[1]{{\color{brown}#1}}

\begin{document}

\begin{frontmatter}

\title{Weakly Supervised Attention-based Models Using Activation Maps \\ for Citrus Mite and Insect Pest Classification}

\author{Edson Bollis, Helena Maia, Helio Pedrini, Sandra Avila}
\address{Institute of Computing, University of Campinas (UNICAMP)\\Campinas, SP, Brazil, 13083-852 }


\ead{edsonbollis@gmail.com, \{helena.maia, helio, sandra\}@ic.unicamp.br}


\begin{abstract}

Citrus juices and fruits are commodities with great economic potential in the international market, but productivity losses caused by mites and other pests are still far from being a good mark. Despite the integrated pest mechanical aspect, only a few works on automatic classification have handled images with orange mite characteristics, which means tiny and noisy regions of interest. On the computational side, attention-based models have gained prominence in deep learning research, and, along with weakly supervised learning algorithms, they have improved tasks performed with some label restrictions. In agronomic research of pests and diseases, these techniques can improve classification performance while pointing out the location of mites and insects without specific labels, reducing deep learning development costs related to generating bounding boxes. In this context, this work proposes an attention-based activation map approach developed to improve the classification of tiny regions called Two-Weighted Activation Mapping, which also produces locations using feature map scores learned from class labels. We apply our method in a two-stage network process called Attention-based Multiple Instance Learning Guided by Saliency Maps. We analyze the proposed approach in two challenging datasets, the Citrus Pest Benchmark, which was captured directly in the field using magnifying glasses, and the Insect Pest, a large pest image benchmark. In addition, we evaluate and compare our models with weakly supervised methods, such as Attention-based Deep MIL and WILDCAT. The results show that our classifier is superior to literature methods that use tiny regions in their classification tasks, surpassing them in all scenarios by at least 16 percentage points. Moreover, our approach infers bounding box locations for salient insects, even training without any location labels.

\end{abstract}

\begin{keyword}
Weakly supervised learning, attention, activation maps, multiple instance learning, insect pests.
\MSC[2021] 00-01\sep  99-00
\end{keyword}

\end{frontmatter}


\section{Introduction}
\label{sec:introduction}

The global consumption of fruit-based drinks reached 95.69 billion liters in 2018 and, among the flavors preferred by consumers in the 100\% juice category, orange is the preferred one, representing 43.8\% of the market~\citep{Neves2020}. The more than 45,763 million tons of sweet orange produced per year in 2016 demonstrate the billionaire global market~\citep{Spreen2020}. However, the productivity of citrus has not reached its full potential and has decreased in the largest orange producers, such as Brazil and the United States, mainly because of hazards caused by pests and diseases~\citep{Spreen2020,Bassanezi2019, DeCarvalho2019}.

\begin{figure}[t]
\captionsetup[subfloat]{farskip=2pt,captionskip=2pt}
\centering
\subfloat[Blur]{\includegraphics[height=1.8cm,width=1.8cm]{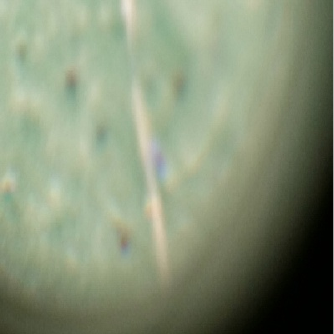}\label{pics:blurred-a}\hspace{0.01cm}
\includegraphics[height=1.8cm,width=1.8cm]{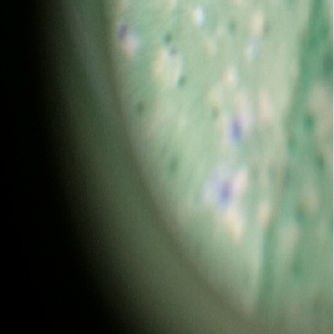}\hspace{0.01cm}
\includegraphics[height=1.8cm,width=1.8cm]{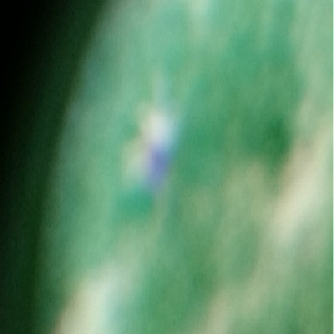}\hspace{0.01cm} 
\includegraphics[height=1.8cm,width=1.8cm]{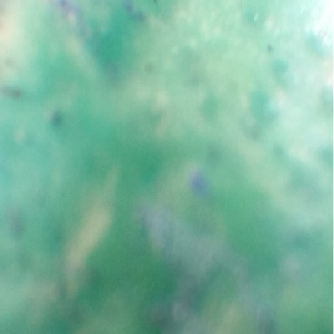}
}  \\

\subfloat[Brightness]{\includegraphics[height=1.8cm,width=1.8cm]{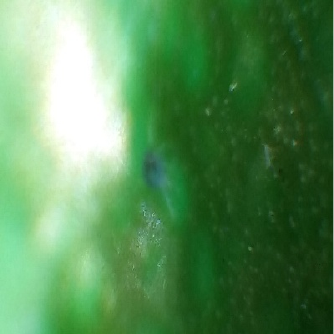}\label{pics:blurred-e}\hspace{0.01cm}
\includegraphics[height=1.8cm,width=1.8cm]{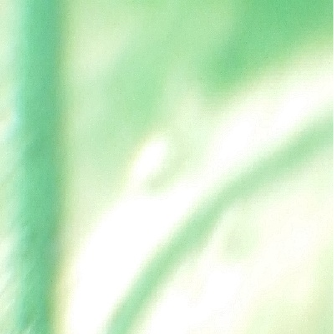}\hspace{0.01cm}
\includegraphics[height=1.8cm,width=1.8cm]{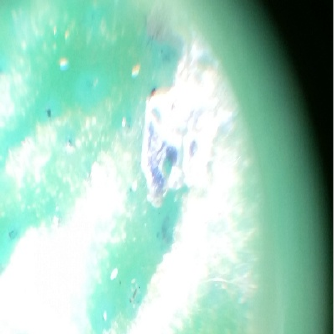}\hspace{0.01cm}
\includegraphics[height=1.8cm,width=1.8cm]{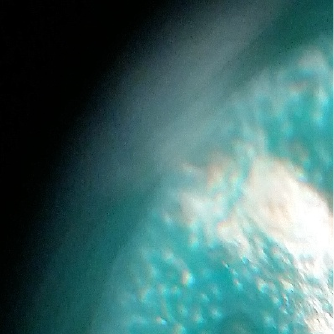}
}\\

\subfloat[No noise]{\includegraphics[height=1.8cm,width=1.8cm]{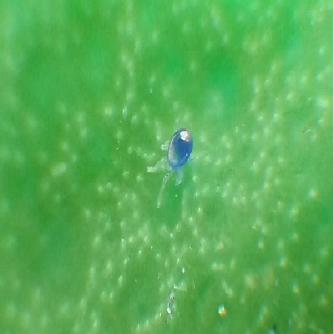}\label{pics:blurred-i}\hspace{0.01cm} 
\includegraphics[height=1.8cm,width=1.8cm]{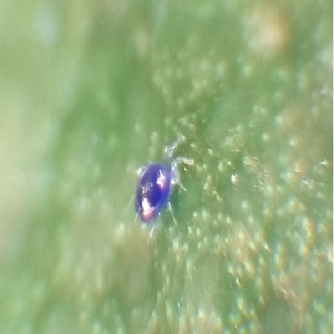}\label{pics:blurred-j}\hspace{0.01cm}  
\includegraphics[height=1.8cm,width=1.8cm]{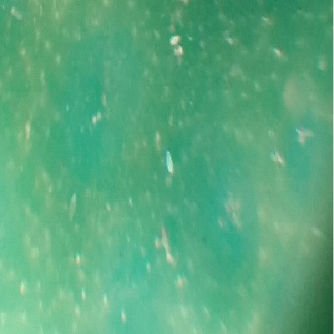}\label{pics:blurred-k}\hspace{0.01cm}
\includegraphics[height=1.8cm,width=1.8cm]{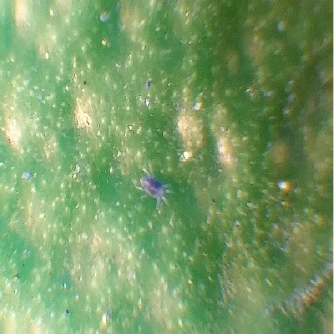}\label{pics:blurred-l}
}  

\caption{
Patches from CPB mites:  each capture uses 60× magnification, each division is 1/9 of the original square size. Mite colors in original images are near the red range, but the range changes after the MIL-Guided cut process in patches. (a) Blur. (b) Brightness. (c)~No~noise.}
\label{pics:blurred}
\end{figure}

Citrus mite control spends more than US\$~54 million a year, a total cost of 5\% in orchard management~\citep{Bassanezi2019}. Most mites are usually invisible to the naked eye. Therefore, finding and classifying their presence is meticulous and costly work. Its process includes magnifying glasses as the primary tool to visualize mites, but, in some cases, mites present a small body area almost imperceptible even with magnification. This is the case of the leprosis vector \textit{Brevipalpus phoenicis}, one of the main threats to citrus orchards, popularly known as leprosis mite or false spider mite (Figures~\ref{pics:blurred-l}~and~\ref{pics:pest-vector-symptom-b}). 

Despite recent abounds of new benchmarks on agronomic pests and diseases~\citep{wu2019ip102, pei2020enhancing, Wang2021}, most threats are regional and mite presence is rare. Datasets often contain images from the Internet, such as the Insect Pest (IP102)~\citep{wu2019ip102}. In contrast, samples collected directly in the field under natural conditions are extremely difficult to obtain, especially properly annotated ones. An available dataset with these characteristics is the Citrus Pest Benchmark (CPB)~\citep{bollis2020weakly}, which contains mite images. Some pest species are more common in a crop from a climatic region, bringing economic impacts in areas equivalent to the size of provinces and states. Distinguishing mites may be a difficult task due to the species variety and the regional aspect. However, identifying their presence is more feasible thanks to the similar~characteristics. 

Applying deep learning techniques demands many labeled images, such as bounding boxes and pixel-level classification. For this reason, the benchmarks available in the literature may be insufficient to train robust models. In addition, creating new complex datasets is an obstacle due to high costs. A good alternative is weakly supervised learning (WSL), which decreases label necessities and development outgoings to surpass these challenges. WSL methods such as activation maps~\citep{zhou2016learning} and multiple instance learning (MIL)~\citep{Dietterich1997} use only classification labels and provide a good level of confidence in localization and classification tasks. In medicine, the use of WSL is common~\citep{durand2017wildcat,ilse2018attention}, but in pest and disease management, just a few works have explicitly proposed the use of WSL~(Section~\ref{sec:wsl-rel}). Early results in citrus mite classification show the WSL classification performance in finding regions of interest (ROIs) with mite characteristics~\citep{bollis2020weakly}, for example, tiny and noisy regions (Figure~\ref{pics:blurred}). 

Recent WSL and deep learning approaches use attention-based models to improve results and understand how models make their predictions~\citep{chen2020robust,yeh2020enhanced}. Attention-based models use attention to highlight multiple features extracted in training time. In the same sense, we propose a new attention-based strategy using two weights to highlight features. Influenced by spatial attention and detection module ideas~\citep{Woo2018,shen2019globally}, our proposed method called Two-Weighted Activation Mapping (Two-WAM) aims to find relevant feature maps in a convolutional neural network (CNN)~\citep{krizhevsky2012imagenet,szegedy2015going} and highlight them through an optimizing process based on feature fusions. Two-WAM forces models to predict a greater number of ROIs, better adapting the inferences on those regions, in consequence, the method produces more reliable locations. We apply Two-WAM in a WSL process called Multiple Instance Learning Guided by Saliency Maps (MIL-Guided)~\citep{bollis2020weakly}, our previous work. 

Our Attention-based Multiple Instance Learning Guided by Saliency Maps (Attention-based MIL-Guided) process successfully classifies whether mites are present on images using its weakly supervised localization produced by Two-WAM, only with class labels. Attention-based MIL-Guided uses Two-WAN localization to simulate a zoom process in mite locations and uses those regions to classify better. Including an attention-based approach as part of MIL-Guided simultaneously allowed the WSL process to improve weakly supervised localization, create activation maps as outputs, and increase feature scores referring to mite areas.

Moreover, as experiments, we analyze Attention-based MIL-Guided and MIL-Guided influence on models trained using Citrus Pest Benchmark (CPB)~\citep{bollis2020weakly} images and patches from these images. Furthermore, we evaluate the impact of noisy regions in models from both processes. For this, the CPB underwent a noisy image removal, creating a Noiseless CPB (NCPB). To evaluate the attention classification performance, we compare the proposed method with the MIL-Guided~\citep{bollis2020weakly}, Attention-based Deep MIL~\citep{ilse2018attention}, and WILDCAT~\citep{durand2017wildcat} using CPB and IP102 datasets. In addition, we perform an ablation study to show the best setup for Attention-based MIL-Guided. 
 
Our key contributions are four-fold:  

\begin{enumerate}

\item We proposed a process called Attention-based Multiple Instance Learning Guided by Saliency Maps (Attention-based MIL-Guided) to classify small regions of interest through the sequential interaction of activation maps and multiple instance learning;

\item We introduced a mathematical formulation method called Two-Weighted Activation Mapping (Two-WAM), which improves the classification of tiny regions, also producing locations using feature map scores learned from class labels.

\item We performed an analysis of the noise influence on deep learning models for tiny and salient regions. We observed that salient image noise positively impacts the models’ classification performance. 

\item We surpassed literature methods (Attention-based Deep MIL and WILDCAT) in all scenarios by at least 16 percentage points on CPB dataset and 1.9~percentage points on IP102 dataset.
\end{enumerate}  

This text is organized as follows. Section~\ref{sec:background} reviews some relevant concepts related to disease and pest classification, weakly supervised learning, multiple instance learning, activation maps, and attention-based methods. Section~\ref{sec:related-work} describes relevant weakly supervised learning approaches and attention-based models available in the literature. Section~\ref{sec:datasets} presents benchmarks used in the experiments. Section~\ref{sec:methods} introduces the Attention-based MIL-Guided process and Two-WAM. We report and discuss the experimental results in Section~\ref{sec:experiments-results}. Finally, some conclusions and directions for future work are presented in Section~\ref{sec:conclusions}.

\section{Background}
\label{sec:background}

This section reviews some concepts related to this work. In Subsection~\ref{sec:ipm}, we overview integrated pest management, pest, and disease vector. Weakly supervised learning is described in Subsection~\ref{sec:wsl}, more specifically, inexact supervision. We also explain relevant aspects of multiple instance learning, activation maps, and saliency maps. In Subsection~\ref{sec:attention-based}, we describe approaches based on attention mechanisms.

\subsection{Integrated Pest Management}
\label{sec:ipm}

Integrated pest management (IPM)~\citep{elliott1995integrated} involves the use of a combination of pest management tactics to reduce economic losses caused by pests to tolerable levels, with minimal environmental side effects. IPM received its first definition in the 1950s~\citep{smith1966}. The IPM describes how to avoid the problems and what are the rules to apply inputs before the problem occurs~\citep{ifas2020Production}. Usually, human inspectors walk along the orchards streets collecting samples to analyze them and reporting the results in paper sheets or mobile tools for data acquisition~\citep{ifas2020Production}. The inspectors examine stalks, leaves, and fruits for hours, trying to find mites and insects to quantify them. The process is mechanical and can be done by machines. In addition, as expected, when humans perform the task, the IPM process is prone to errors due to the inability or fatigue of the handlers~\citep{bollis2020weakly}.

\begin{figure}[tb]
\centering
\subfloat[Rust mites~\citep{wu2019ip102}]{
\includegraphics[height=3.0cm,width=3.6cm]{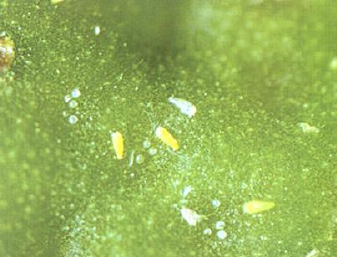} \label{pics:pest-vector-symptom-a}
}  \hspace{.01cm}
\subfloat[False spider mite~\citep{bollis2020weakly}]{
\includegraphics[height=3.0cm,width=3.6cm]{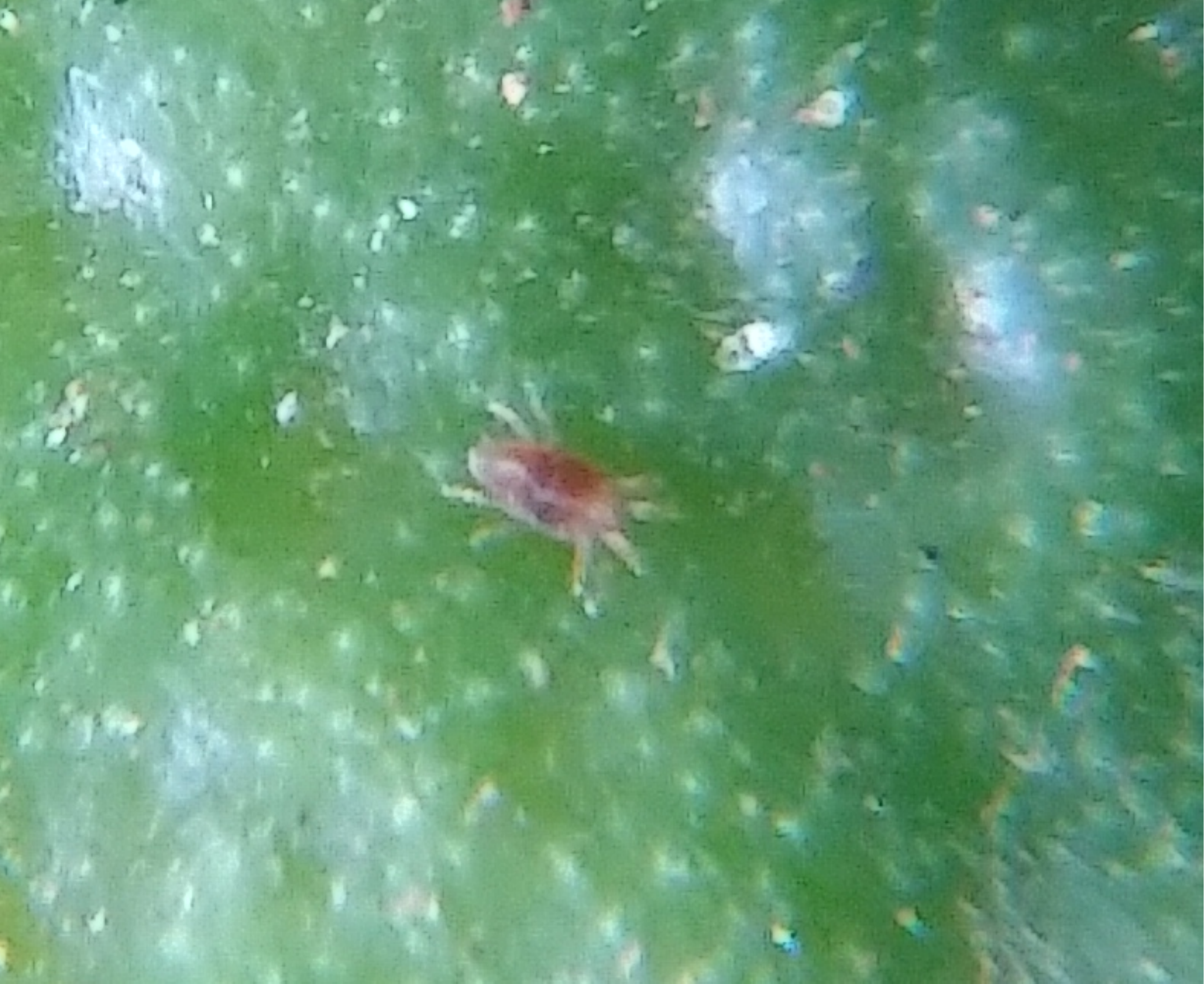} \label{pics:pest-vector-symptom-b}
}  \\

\subfloat[Rust mite symptoms~\citep{wu2019ip102}]{
\includegraphics[height=3.0cm,width=3.6cm]{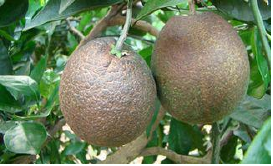} \label{pics:pest-vector-symptom-c}
}  \hspace{.01cm}
\subfloat[Leprosis virus symptoms~\citep{bastianel2006citrus}]{
\includegraphics[height=3.0cm,width=3.6cm]{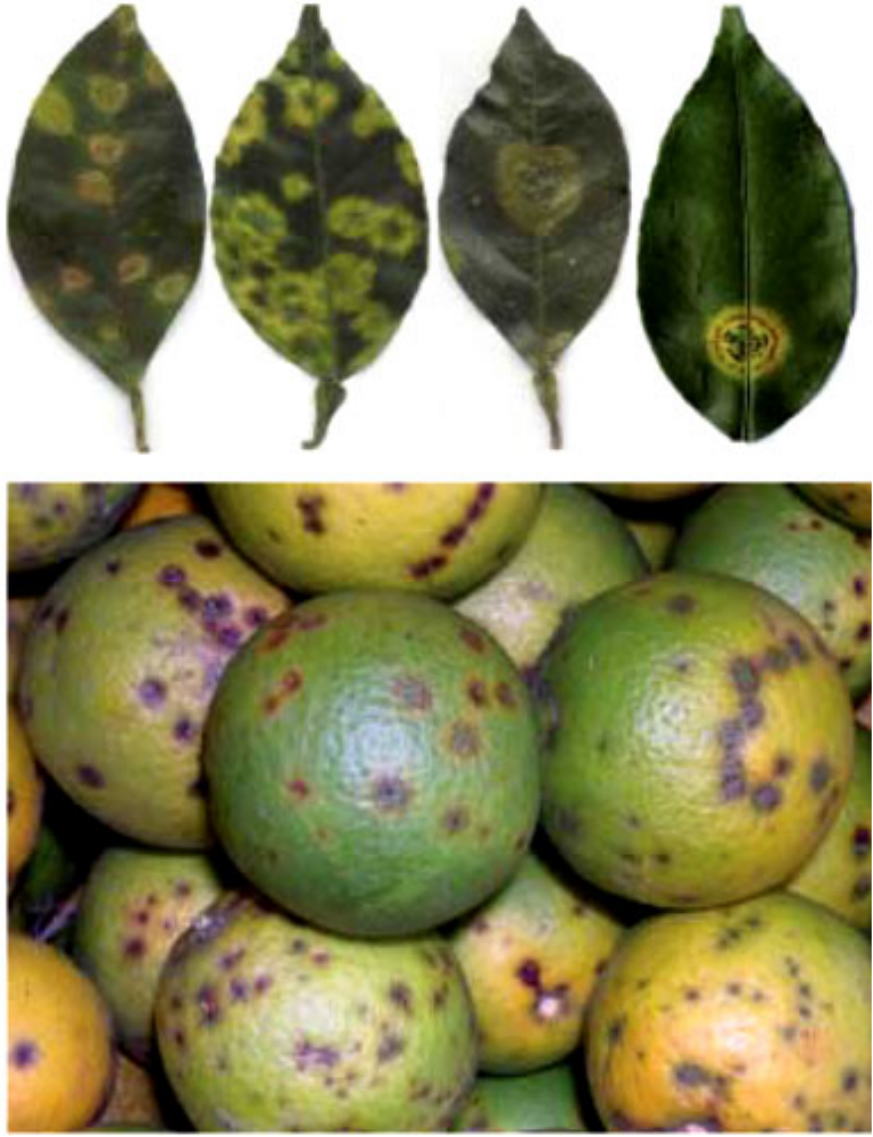} \label{pics:pest-vector-symptom-d} 
}  \\

\caption{(a) Rust mites (\textit{Phyllocoptruta oleivora}). (b) False spider mite  (\textit{Brevipalpus phoenicis}) (c) Symptoms produced by rust mites. (d) Symptoms produced by leprosis viruses transported by a false spider mite. Images reproduced from~\citep{wu2019ip102, bollis2020weakly, bastianel2006citrus}.}
\label{pics:pest-vector-symptom}
\end{figure}

In the IPM automation literature, works are typically divided into two groups. Pest-related studies~\citep{Woo2018,Wang2020a} deal with insects or mites that cause crop losses, i.e., they are the reason for production losses and the appearance of symptoms, such as rust mites in Figure~\ref{pics:pest-vector-symptom-a} and false spider mites in Figure~\ref{pics:pest-vector-symptom-b}. Symptom-related tasks~\citep{Lu2017a, Cap2020} deal with damages that have already occurred to leaves, stems, and fruits, such as the rust mite symptom in Figure~\ref{pics:pest-vector-symptom-c} and leprosis virus symptom in Figure~\ref{pics:pest-vector-symptom-d}.    

\subsection{Weakly Supervised Learning}
\label{sec:wsl}

Weakly supervised learning (WSL) is an umbrella term to refer to methods for training models using weak or unreliable labeled data.  Typically, there are three types of weak supervision~\citep{zhou2018brief}: incomplete, inaccurate, and inexact. Incomplete supervision assumes two sets of training data. One set is usually smaller and labeled, and the other unlabeled. Inaccurate supervision uses only one set of unreliable labels, i.e., labels may contain mistakes and, as a consequence, a subset is partially or totally wrong. Inexact supervision presents a unique set provided with coarse-grained labels. It is expensive to spend resources on hiring technicians to analyze and generate thousands of small bounding boxes in our context. Our datasets only contain class labels for mites, and even using class labels, we intend to find mites to use their location in our process. In this way, we model our method as an inexact supervised task or, more specifically, based on a combination of activation maps and MIL methods.

\subsubsection{Multiple Instance Learning} 
\label{sec:MIL}

Multiple instance learning (MIL)~\citep{Dietterich1997, Carbonneau2018} is a form of weakly supervised learning in which the data is arranged in sets of instances called bags. Tasks related to MIL fall into two distinct objectives: whole bags, which is the most common, and individual instances, which typically is an intermediate step. The bag inference consists of using several instances to calculate the final prediction at once. Instance inference predicts a result for each instance and then aggregates those results into a final prediction for the bag.

In training time, we assume that instances receive the same labels as its bags, i.e., $g(x_{ij}) = f(X_i) = y_i$, where $X_i$ is the $i$-th bag, $x_{ij}$ its $j$-th instance and $g$ is a function that maps each instance from $\{x_{ij}, i \in \mathbb{N}, j \in \mathbb{N}\}$ into a label from the set $Y$. In this work, we define a bag $X_i$ as a set of parts or patches $x_{ij}$ extracted from an image $x_i \in \mathbb{R}^{W \times H \times 3}$ from the original dataset.
Based on these concepts, a task in MIL consists of learning a model $F: X \rightarrow Y$ for a function $f: X \rightarrow Y$ using a training dataset $D_s =\{(X_i,y_i), f(X_i) = y_i, X_i = \{x_{ij}, j = 1,..., m_i \in \mathbb{N}\} \subset X, i = 1,...,n \in \mathbb{N}\}$ to predict labels for a test set $D_t$. The value $n$ is the total number of bags and $m_i$ the number of instances in each bag $i < n$.

\subsubsection{Activation Maps and Saliency Maps}
\label{sec:act_map}

Considering an image $x \in X$, saliency maps are models or functions defined as $S: X \rightarrow [0,1]^{W \times H}$ where $X \subset \mathbb{R}^{W \times H \times 3}$, and $W$ and $H$ are width and height in pixels. Saliency map values close to 1 show ROIs, and close to 0 represent not interesting areas. It is common to exhibit $S(x)$ in thermal images, where red represents values close to 1, yellow intermediary values, and blue values close to 0.

Activation maps are saliency maps generated by convolutional neural networks (CNN)~\citep{zhou2016learning}. Activation maps ($M_{\textit{act}} \in \mathbb{R}^{w \times h} $ in Equation~\ref{eq:cam}, with $w,h \in \mathbb{N}$) are maps from images $x \in X \subset \mathbb{R}^{W \times H \times 3}$, where $f_k$ is the k-th feature map (i.e. a matrix produced by a convolutional model), $w_k \in \mathbb{R}$ is the weight of each $k$-th feature map, and $A_f$ is an activation function. For visualizing $M_{\textit{act}}$ jointly with an original image, it is necessary to scale $M_{\textit{act}}$ until $x$ size ($h,w$ to $H,W$) and normalize the elements from the resultant matrix to values between $[0,1]$.

Different activation map methods find $w_k$ in different ways, and they make $w_k$ class dependent. Some methods modify input images to alter the classifier behavior and use the regions of modifications as the saliency maps~\citep{Wang2020a}. In this work, we use the term activation maps to refer to $M_{\textit{act}}$, and saliency maps to refer to the final result after scaling and normalizing the matrix.
\begin{equation}
\label{eq:cam}
M_{\textit{act}}(x) = A_f\left(\sum_{k} w_k \cdot f_k(x)\right).
\end{equation}
    
\subsection{Attention-based Approaches and Feature Selection} 
\label{sec:attention-based}

In deep learning architectures, attention-based approaches produce attention-based models~\citep{bahdanau2014neural,raffel2015feed}. Attention-based models incorporate relevance notion by allowing the model to dynamically focus attention to only certain parts of the input that effectively perform the task~\citep{Chaudhari2019a}. It means that they select relevant features, which improve and support model decisions.

Some authors distinguish weakly supervised feature selectors from attention-based approaches for including some structures on the selection process, while it is implicit in attention-based~\citep{durand2017wildcat}. However, internally, both concepts appeared from biological visual attention~\citep{Borji2019}. Most attention structures apply WSL selection strategies to highlight or hide features~\citep{Choe2020}, making their concepts very close to the activation maps.

Attention-based approaches typically use weight optimization, and high attention weights directly correspond to ROIs~\citep{adiga2020manifold}. In this work, attention-based activation maps produce saliency maps based on the optimization weights trained for this purpose. Equation~\ref{eq:mask} shows a usual formulation for producing attention-based activation maps in which $f_{k}$ and $M_{\textit{act}}$ are the same elements from  Equation~\ref{eq:cam} (it is common to use $\otimes$ as an element-wise multiplication). \begin{equation} \label{eq:mask}
f^{\otimes}_{k}(x) =  f_k(x) \otimes M_{\textit{act}}(x).
\end{equation}

\section{Related Work}
\label{sec:related-work}

This section reviews the methods directly relevant to this work: weakly supervised methods (Subsection~\ref{sec:wsl-rel}) and attention-based methods (Subsection~\ref{sec:attention-rel}). We also select works proposed to address agriculture problems for each section, particularly for image-based pest and disease classification. For a comprehensive review, we refer the reader to the survey by~\citet{Rony2019,Wang2019c} about weakly supervised learning methods and~\citet{Correia2021} about attention-based methods.

\subsection{Weakly Supervised Learning}
\label{sec:wsl-rel}

\noindent \textit{General Methods:} 
MIL has been widely applied in many areas of machine learning. \citet{ilse2018attention} proposed a general procedure for modeling the bag label probability and incorporating interpretability to the MIL approach. Attention-based Deep Multiple Instance Learning (Attention-based Deep MIL) presented a weighted average of instances (low-dimensional embeddings), where a neural network determines weights. The attention weights called the gated attention mechanism allow the method to find key instances, which were used to highlight ROIs, as illustrated in Figures~\ref{pics:activation-maps-b} and~\ref{pics:ip102-activation-maps-b}. Attention-based Deep MIL was evaluated on several image classification datasets, including five MIL benchmarks, an MNIST-based image dataset, and two histopathology datasets. The inputs to the Attention-based Deep MIL model are the cells of a grid from a dataset image (see Figure~\ref{pics:activation-maps-b}). These inputs receive the same label as the corresponding image, even regions that do not contain mites. In contrast, our proposed method uses activation maps to guide the patch extraction to ensure that only the regions of interest are used in the network.

Activation maps have emerged as an attempt to explain a CNN~\citep{Rony2019}. \citet{zhou2016learning} proposed the Class Activation Mapping (CAM) technique or commonly known as activation maps (Subsection~\ref{sec:act_map}). Inspired by this work, \citet{selvaraju2017grad} created Gradient-weighted Class Activation Mapping (Grad-CAM), which uses the gradients of any target concept, flowing into the final convolutional layer to produce a coarse localization map highlighting the important regions in the image for predicting the concept. Although Grad-CAM can locate relevant regions, the gradient computation may be expensive in restricted devices such as mobiles. This type of handheld device would be particularly useful for inferences in orchards. By introducing the Two-WAM in our proposed model, the network is able to directly infer activation maps without further steps, enabling its use in restricted devices.

\citet{durand2017wildcat} proposed a weakly supervised method for multi-label object classification, localization, and segmentation called WILDCAT (Weakly supervIsed Learning of Deep Convolutional neurAl neTworks). They applied a CNN as a feature extractor and, instead of pooling information only from the maximum scoring, WILDCAT includes the minimum scoring regions to regularize the class score. Moreover, as WILDCAT does not use a fully connected layer, feature maps are combined separately to generate class-specific heatmaps that can be globally pooled to get a single probability for each class. They call this strategy class-wise pooling. WILDCAT was evaluated on various visual tasks, such as image classification, object recognition, and scene categorization. WILDCAT, as well as Grad-CAM, present great results in the location and classification of salient regions, and thus may be suitable for general pest detection. However, they are not focused on tiny regions, an important aspect of mites. Besides the bag model, we need to employ an instance model to capture fine-grained structures to address this issue. 

In our experiments, we compare our approach to Attention-based Deep MIL and WILDCAT methods.\vspace{0.1cm}

\noindent \textit{Pest and Disease Classification Methods:} 
In the agriculture field, \citet{Lu2017a} proposed the first weakly supervised method for identifying crop disease symptoms~\citep{Kim2020}. Their wheat disease diagnosis system identifies disease categories and localizes corresponding disease areas simultaneously for in-field wheat images. They applied a fully convolutional VGG~\citep{simonyan2014very} to extract local features from the whole image and generate spatial score maps, where each score point is a disease estimation for the corresponding receptive field. They treated the image whose receptive fields cover salient objects as a positive bag for its corresponding class label in MIL. 

\citet{bollis2020weakly}'s work is the first to apply weakly supervised learning for pest classification. The method was designed as a weakly supervised multiple instance learning method guided by saliency maps to automatically select ROI in mite images. We detail this method, the Multiple Instance Learning Guided by Saliency Maps (MIL-Guided), in Subsection~\ref{sec:MIL-Guided} and its extension, Attention-based Multiple Instance Learning Guided by Saliency Maps (Attention-based MIL-Guided), in Subsection~\ref{sec:mil-masking}.

\citet{Chen2021a} developed a fine-grained fly species classification (FGFSC) method. Some fly species images came from the IP102~\citep{wu2019ip102} dataset, and others came from image search engines. The FGFSC first obtains and extracts regions of the image that contains a fly to decrease the background influence. Next, the FGFSC learns how to choose sub-areas of the fly region to obtain local embedding and concatenate them with the global embedding. The classifier then considers refined characteristics to better distinguish among fly species. As MIL-Guided~\citep{bollis2020weakly}, they first extract patches from original images using Grad-CAM~\citep{selvaraju2017grad}, and they then use glimpses of regions to train a feature fusion network for dealing with local characteristics.

\citet{Wang2020a} proposed a weakly supervised localization method based on multiscale saliency maps fusion for classifying mangrove pest images. They also created a forestry pest image dataset called Mangrove Insect Pest Dataset in Guangxi, China (MIPDGC), mainly for mangrove wetland ecosystems. The dataset, which is not available, containing more than 50,000 images belonging to 120 categories of forest~pests. 

\citet{Cap2020} introduced a weakly supervised leaf segmentation method (LFLSeg) that helps the classification model to learn the dense and interior leaf regions implicitly. They obtained the segmented leaf region by employing the Grad-CAM technique \citep{selvaraju2017grad}. 
LFLSeg is used to guide the proposed LeafGAN, a generative adversarial networks (GAN) based data augmentation method, to focus on the leaf regions for generating images of cucumber disease plants. The authors collected 19,431 cucumber leaf images, which are not available, from multiple locations in Japan. 

It is noteworthy that strategies in the IPM automation literature consider images with visually salient ROIs~\citep{Wang2020a,Chen2021a}. On the other hand, dealing with images containing tiny ROIs is challenging. New works have studied tiny insects. For example, the following works detected and localized aphids in wild conditions~\citep{pei2020enhancing, Wang2021}, and counted and classified aphids in controlled environments~\citep{lins2020method}. Aphids are small but visible, unlike some species of mites, which are unrecognizable without magnification. Moreover, some works dealt only with disease symptoms~\citep{Lu2017a, Wu2019c}. For IPM, symptoms and pests are essential, and preventing pests helps improve production while dealing with symptoms to mitigate losses.

\subsection{Attention-based Models}
\label{sec:attention-rel}

\noindent \textit{General Methods:} 
In 2014, attention-based approaches became a fundamental concept in neural networks and state of the art in many tasks~\citep{Correia2021}. EfficientNet-B0~\citep{tan2019efficientnet}, the backbone chosen in this work, is composed of mobile inverted bottleneck convolution blocks~\citep{Howard2019} and squeeze-and-excitation (SE) attention layers~\citet{Hu2020}. This attention layer adaptively recalibrates channel-wise feature responses by modeling interdependencies between channels.

\citet{Woo2018} explored both spatial and channel-wise attention. To compute the channel attention, they squeezed the spatial dimension of the input feature map, and to compute the spatial attention, they applied average-pooling and max-pooling operations along the channel axis and concatenated them to generate a feature descriptor. The channel attention focuses on `what' is meaningful given an input image, while the spatial attention focuses on `where' is an informative part.

Towards spatial attention, \citet{shen2019globally} employed the SE strategy in a MIL-based approach, similar to MIL-Guided. However, they trained the part of their model considering the entire image and generated predictions for instances in an end-to-end pipeline. In this way, the second part influences the first part in training time, which may be negative. The evidence came from the average predictions of the two parties when separated. The model for instances was not much better in its predictions than the results for the entire image model. For this reason, our Attention-based MIL-Guided trains models separately and the Instance Model reached much better results in comparison with the Bag Model (Subsection \ref{sec:comparison-CPB}). In Attention-based Deep MIL, \citet{ilse2018attention} applied attention in a feature vector group to substitute a pooling layer based on vectors. This attention pooling method allows the instances to contribute more with the classification when ROI appears in images. It is also a form of spatial attention.\vspace{0.1cm}

\noindent \textit{Pest and Disease Classification Methods:}
\citet{Liu2019c} proposed an attention module following the same spatial idea from~\citet{Woo2018}. Instead of using weights to generate activation maps directly, they employed layers in sequence to produce activation maps. First, they applied a point-wise convolution to compress the feature maps in only one after using a 7$\times$7 kernel followed by a transposed convolution and a deep-wise multiplication (Equation~\ref{eq:mask}). Their method highlights insects in pest image acquisition equipment (traps) for multi-class detection and classification. 

\citet{Wang2020} explored a channel attention mechanism from~\citet{Hu2020} in each projection convolution block and residual block to address in-field pest detection and counting. They introduced the In-Field Pest in Food Crop (IPFC) dataset (not available), which contains 17,192 in-field pest images.

\citet{Zeng2020} introduced a self-attention network for creating and selecting feature maps in symptom classification. They used matrices as values, keys, and queries produced from convolutions applied to feature maps. \citet{Deng2018a} and \citet{Nanni2020} utilized saliency maps as attention to original pest images, hiding background regions. They used a dataset proposed by~\citet{Deng2018a} composed of ten different pests found mainly on tea plants and other plants spread between Europe and Central Asia. In addition to the original images, they used activation maps as inputs to improve the classification effectiveness\HM{. }
    
Unfortunately, most agriculture works do not share their source code. In some cases, codes are available but only for specific tasks such as localization and image generation. Therefore, in this work, we compare our results with weakly supervised established methods, in which codes are made freely available and extensively used in literature. We plan to make our codes available after the publication acceptance.

\section{Datasets}
\label{sec:datasets}

This section describes the datasets used in the experiments. We describe the Citrus Pest Benchmark (Subsection~\ref{sec:citrus-pest-benchmark}), and the Insect Pest dataset (Subsection~\ref{sec:IP102}).

\subsection{Citrus Pest Benchmark}
\label{sec:citrus-pest-benchmark}

Citrus Pest Benchmark\footnote{\url{https://github.com/edsonbollis/Citrus-Pest-Benchmark}} (CPB) consists of 10,816 images of 1,200$\times$1,200 pixels. The CPB contains 7,361 mites divided into six mite common types in Brazilian green belt citriculture: (i) spider mites (\emph{Panonychus citri}, \emph{Eutetranychus banksi}, \emph{Tetranychus mexicanus}), (ii) phytoseiid mites (\emph{Euseius citrifolius}, \emph{Iphiseiodes zuluagai}), (iii) rust mites (\emph{Phyllocoptruta oleivora}), (iv) false spider mites (\emph{Brevipalpus phoenicis}), (v) broad mites (\emph{Polyphagotarsonemus latus}), (vi) two-spotted spider mites (\emph{Tetranychus urticae}). All of them are considered positive images in our binary context. CPB also contains 3,455 negative images from citrus fruits and leaves. The data is split into 6,380 training, 2,239 validation, and 2,197 test images (see Table~\ref{table:benchmarks}). 

The images were collected via a mobile device attached with a 60$\times$ magnifier. Even so, some mite species (e.g., rust and false spider mites) represent a very tiny proportion of the entire image. Also, some images have poor quality due to noise (see Figure~\ref{pics:blurred}). 

In order to evaluate the impact of noise, we manually removed the noisy images from the training and validation sets. We called this dataset Noiseless CPB (NCPB). The criterion used to remove noisy images was based on the visibility of the mites in the images. When we could not identify whether there were mites in the images, we removed them. Concerning the negative class, we removed images in which their noisy regions could be mites. NCPB training set decreased to 3,243 positive images and 1,532 negative images, totaling 4,775. NCPB validation set reached 1,142 positive images and 524 negative images, totaling 1,666. NCPB contains approximately 75\% of each CPB training/validation set. 

The classification performance is evaluated using accuracy and F1-score metrics.

\subsection{Insect Pest Dataset}
\label{sec:IP102}

Insect Pest dataset (IP102)~\citep{wu2019ip102} consists of 102 classes and 75,222 images collected from the Internet. The data is split into 45,095 training, 7,508 validation, and 22,619 test images for the insect pest classification task. The IP102 is currently at its version~1.1, released after identifying data annotation errors in its previous version. The new version contains the same number of images as its old version. We refer to IP102's new version as IP102.

IP102 has a hierarchical structure and each super-class assigns sub-classes according to the type of damaged crops: field (e.g., rice, corn, wheat, beet, and alfalfa) and economic (e.g., mango, citrus, and vitis). Despite its name, the IP102 dataset contains mite classes such as winter grain mites or blue oat mites (\emph{Penthaleus major}), red mites (family \emph{Tetranychidae}), and rust mites (\emph{Phyllocoptruta oleivora}). Mites are not insects but belong to the related biological class \emph{Arachnida}~\citep{Costello2000}.

For the experiments conducted on IP102, we resized all images to 224$\times$224 pixels.
The classification performance is evaluated using the standard metrics for the IP102, accuracy, and F1-score.

\begin{table}[!htb]
\setlength{\tabcolsep}{1.0mm}
\begin{center}
\caption{Description of the datasets used in the evaluation of the classification networks.}\label{table:benchmarks}\vspace{0.1cm}
\small
\begin{tabular}{lcrrrr}
\toprule
 Dataset & Classes & Training & Validation & Test & Total \\
\midrule
CPB~\citep{bollis2020weakly} & $2$ & $6,380$ & $2,239$ & $2,197$ & $10,816$ \\
NCPB  & $2$ & $4,775$ & $1,666$ & $-$ & $6,441$ \\
IP102~\citep{wu2019ip102}  & $102$ & $45,095$ & $7,508$  & $22,619$ & $75,222$ \\
\bottomrule
\end{tabular}
\end{center}
\end{table}

\section{Methods}
\label{sec:methods}

This section details the methodology proposed and used in the experiments. We first describe our previous work~\citep{bollis2020weakly}, the so-called Multiple Instance Learning Guided by Saliency Maps (MIL-Guided) (Subsection~\ref{sec:MIL-Guided}). Next, we introduce its extension called Attention-based MIL-Guided (Subsection~\ref{sec:mil-masking}), and our attention-based activation map, the Two-weighted Activation Mapping (Two-WAM) (Subsection~\ref{sec:fea-transf}).

\subsection{Multiple Instance Learning Guided by Saliency Maps}
\label{sec:MIL-Guided}

Multiple Instance Learning Guided by Saliency Maps (MIL-Guided)~\citep{bollis2020weakly} is a WSL process to automatically select ROIs in the images, significantly reducing the annotation task. Our previous method consists of four steps: (1) we train a CNN (initially trained on the ImageNet) on the original dataset $x \in X = \{x_{i}, i=1,\cdots,n\}$ with labels $Y = \{y_{i}, f(x_i)=y_i, i=1,...,n\}$, resulting in the Bag Model; (2) we use a method called Multi-patch Selection Strategy based on Saliency Maps (Patch-SaliMap). Patch-SaliMap generates patches from the original images based on the saliency map scores. It selects the points with the highest scores from saliency map matrices and cuts patches centered on them. These patches become a new dataset $D_s =\{(X_i,y_i), f(X_i) = y_i, X_i = \{x_{ij}, j < m_i = m \in \mathbb{N}\}\}$, where $m$ is constant and represents the number of instances (patches); (3) we fine-tune the Bag Model using the patches, resulting in the Instance Model; and (4) we apply a weighted evaluation scheme called Weighted Evaluation Method to generate a prediction $P$ for the bag (entire image $x_i$) using the prediction $p$ for each patch~$x_{ij}$ (see Equation~\ref{eq:hem}).
\begin{equation}
\label{eq:hem}
P(x_i) = \frac{\displaystyle\sum_{j=1}^{m} (m-j+1) \cdot p(x_{ij})}{\displaystyle\sum_{j=1}^{m} (m-j+1)}.
\end{equation}

In summary, the MIL-Guided process turns MIL tasks based on CNNs into fully supervised steps. Initially, it trains the Bag Model as a fully supervised model. The Bag Model provides feature maps to create saliency or activation maps for the Patch-SaliMap. Patch-SaliMap crops original images based on ROIs from saliency maps and creates instances containing ROIs. The Instance Model makes inferences over instances, and we train it as a fully supervised model. Then, the Weighted Evaluation Method uses the instance probabilities produced by the Instance Model and predicts a probability for the whole bag, i.e., the original image.

\subsection{Attention-based Multiple Instance Learning Guided by Saliency Maps}\label{sec:mil-masking}

\begin{figure*}[!htb]
\captionsetup[subfloat]{farskip=2pt,captionskip=2pt}
\centering
\subfloat[Attention-based MIL-Guided]{
\includegraphics[width=1.0 \textwidth,height=0.2\textwidth]{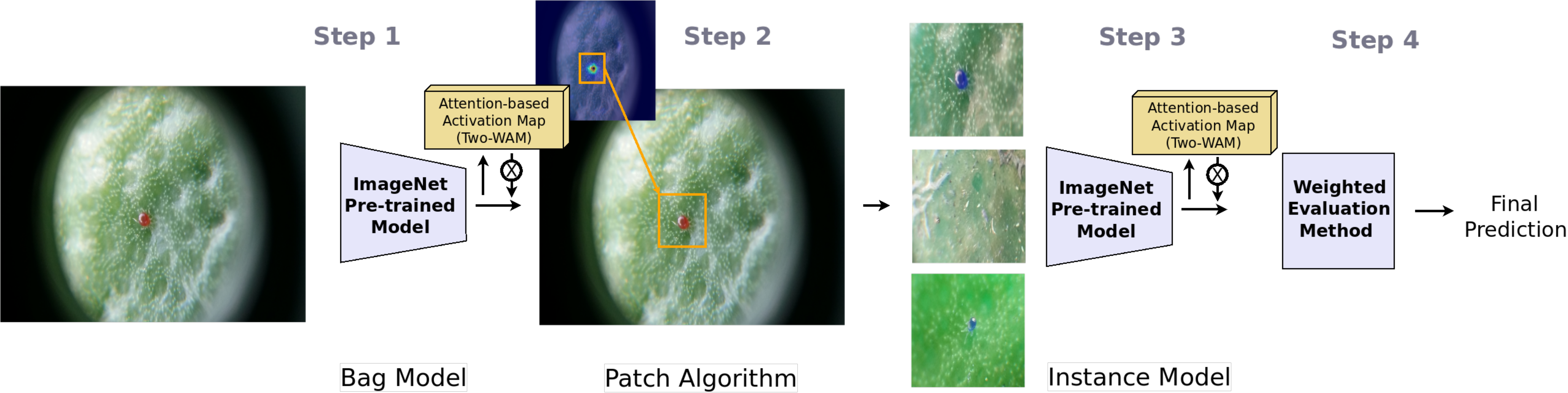}
\label{pics:pipeline-a}} \\
\subfloat[MIL-Guided~\citep{bollis2020weakly} Steps 1 and 2]{
\includegraphics[width=0.36\textwidth,height=0.2\textwidth]{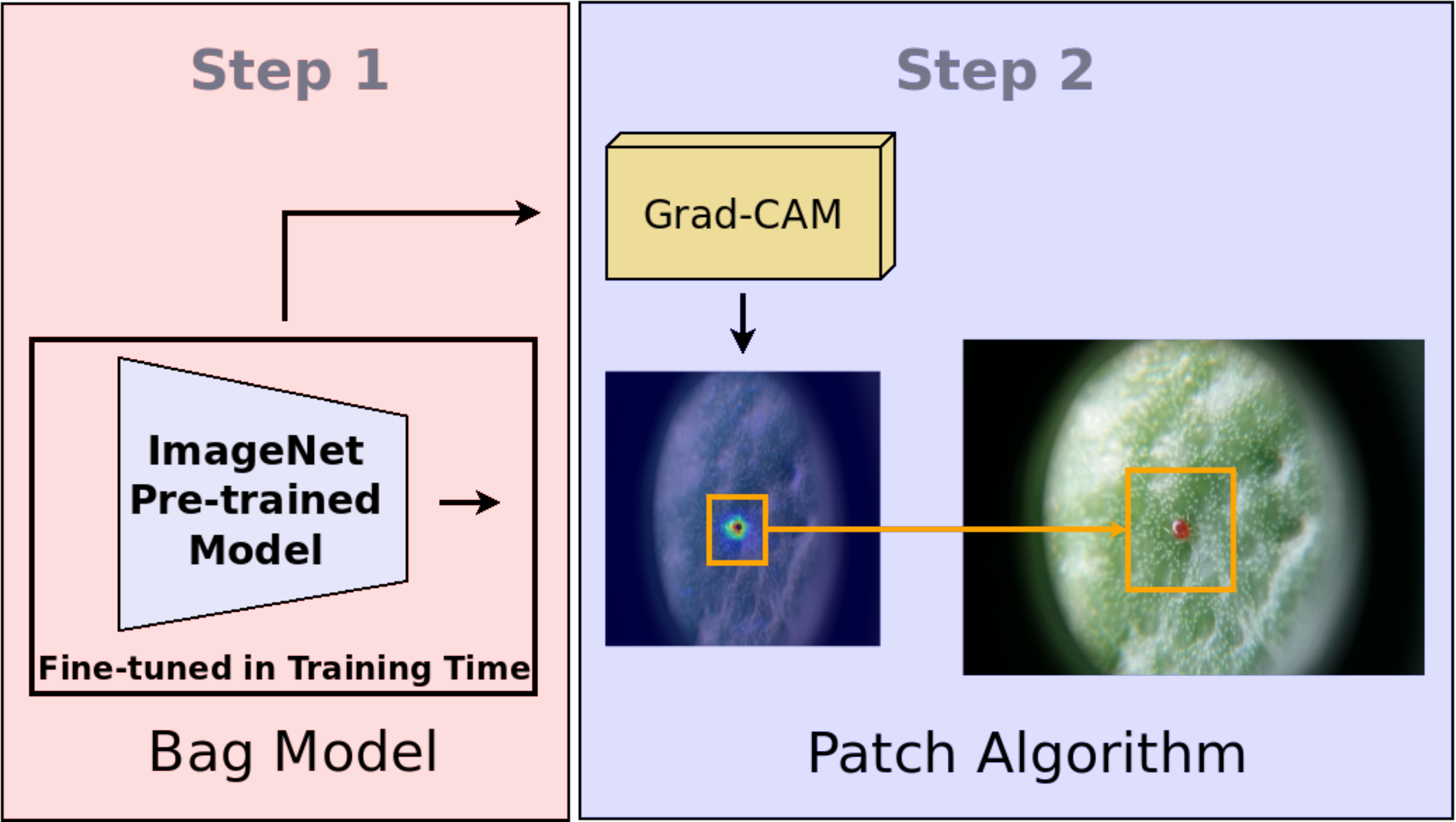} \label{pics:pipeline-b}
} \hfil  \subfloat[Attention-based MIL-Guided Steps 1 and 2]{ 
\includegraphics[width=0.4\textwidth,height=0.2\textwidth]{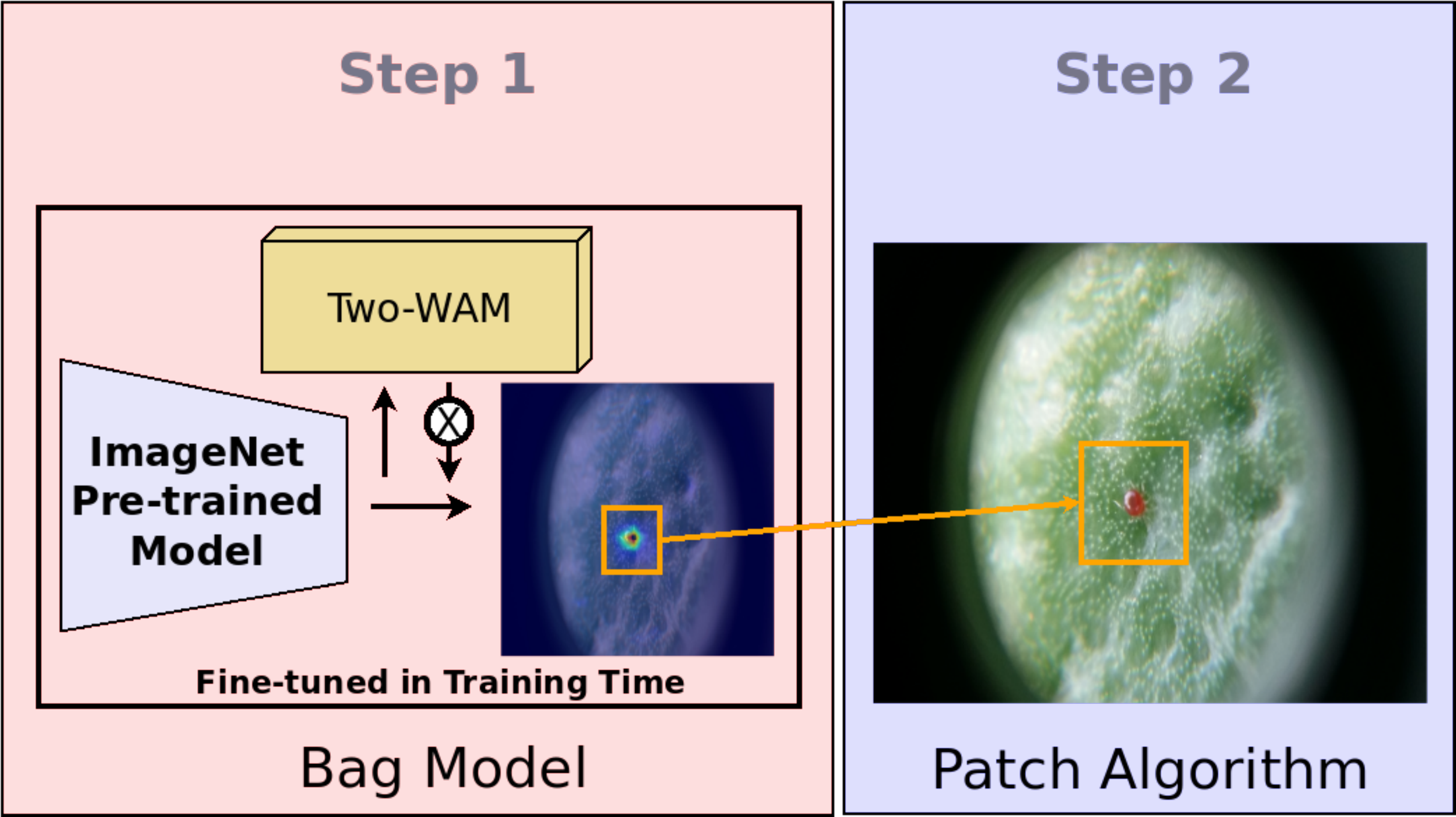} \label{pics:pipeline-c}
}
\caption{(a) Attention-based Multiple Instance Learning Guided by Saliency Maps (Attention-based MIL-Guided) consists of four steps. In Step~1, we train a CNN with an attention-based approach (initially trained on ImageNet), i.e., we create a model for fully supervised classification that also provides weakly supervised localization. In Step 2, we automatically generate multiple patches using the masking-based saliency maps. In Step 3, we train our CNN model (initially trained on ImageNet) according to the multiple instance learning task of inferring over the instances (as a fully supervised classifier). In Step 4, we apply a weighted evaluation scheme to predict the image class. (b) MIL-Guided version of Steps 1 and~2.  In Step 1, we train a CNN, and in Step 2, we generate maps using Grad-CAM. (c) Detailed Attention-Based MIL-Guided version of Steps 1 and 2. The symbol $\otimes$ denotes element-wise multiplication. Figure adapted from Bollis et al.~\citep{bollis2020weakly}.}
\label{pics:pipeline}
\end{figure*}

We introduce Attention-based Multiple Instance Learning Guided by Saliency Maps (Attention-based MIL-Guided), which extended MIL-Guided by exploring attention-based activation maps, as illustrated in Figure~\ref{pics:pipeline}. We take advantage of the attention-based activation map layer to highlight ROI and hide background~features.

Our attention-based approach anticipates the activation map creation, bringing it to the first step. Unlike the MIL-Guided pipeline, the Attention-based MIL-Guided Bag Model directly produces activation maps and, consequently, saliency maps, which are fed to Patch-SaliMap (Figure~\ref{pics:pipeline-c}). MIL-Guided uses Bag Model as input to Grad-CAM, generating the activation maps (Figure~\ref{pics:pipeline-b}). Attention-based MIL-Guided Bag and Instance Models follow the same architecture as MIL-Guided, except for the attention-based activation map layer. Also, the Attention-based MIL-Guided Instance Model fine-tunes an ImageNet pre-trained model instead of the Bag Model (see Subsection~\ref{sec:abalation_study}). We apply the Weighted Evaluation method as MIL-Guided.

The MIL-Guided process is independent of the CNN architecture. Differently, Attention-based MIL-Guided requires an architecture that generates activation maps as outputs. Two-WAM, introduced in the following section, is our approach to generate an activation map method, which learns how to use ROIs to influence the training process and, at the same time, to improve its locations through the learned weights. 

\subsection{Two-Weighted Activation Mapping}
\label{sec:fea-transf}

We define Two-Weighted Activation Mapping, or Two-WAM for short, as a transformation $T: \mathbb{R}^{w \times h \times k} \to \mathbb{R}^{w \times h}$, where $w$, $h$ and $k$ denote the number of rows, columns and feature maps, respectively. In other words, Two-WAM is a method that transforms a group of $k$ feature maps into only one, using two optimized weights for each feature map. The process is similar to the pointwise convolution (1$\times$1 convolution~\citep{szegedy2015going}), except that (i) we use a single scalar multiplication between the weights and a channel matrix instead of using a sliding kernel and (ii) we model the combination of the channels' results as a polynomial function with linear and exponential coefficients (weights).

We formally define the general transformation for our attention-based activation map layer to consider that feature maps are tensors of floating-point values varying in the real $n$-dimensional space. For this reason, we define a transformation among $k$ feature maps, such as Equation~\ref{eq:trans} (which can be rewritten in the form of Equation~\ref{eq:cam}), where $\alpha_k$ and $\beta_k$ are weights that the transformation learns in training time, and $c$ is a constant value. The final result $T_{act}$ is our activation map. $T_{act}$ highlights original feature maps~$f_k$ for  $i = 1,\dots,n$ through the mathematical operation described in Equation~\ref{eq:transf}~(same as Equation~\ref{eq:mask})~\citep{adiga2020manifold}, where $\otimes$ denotes an element-wise multiplication.
\begin{equation}
\label{eq:trans}
T_{act}(x) =  \frac{\sum_{k} \alpha_k \cdot f_k(x) \cdot c^{\beta_k}}{\sum_{k} \alpha_k \cdot c^{\beta_k}}.
\end{equation}
\begin{equation} \label{eq:transf}
f^{\otimes}_{k}(x) =  f_k(x) \otimes T_{act}(x).
\end{equation}

\begin{figure}[!htb]
\captionsetup[subfloat]{farskip=2pt,captionskip=2pt}
\centering
\subfloat[{$[R,G,B]$}]{\includegraphics[height=2.8cm,width=3.8cm]{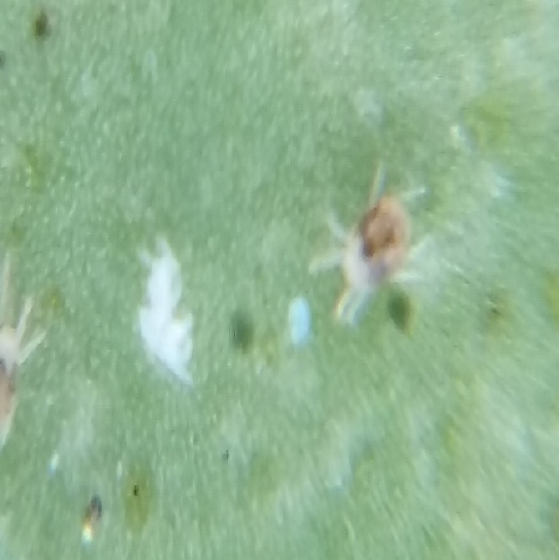}\label{pics:RGB-Transf-a}} \hspace*{0.1cm}
\subfloat[{$T([R,G,B])$}]{\includegraphics[height=2.8cm,width=3.8cm]{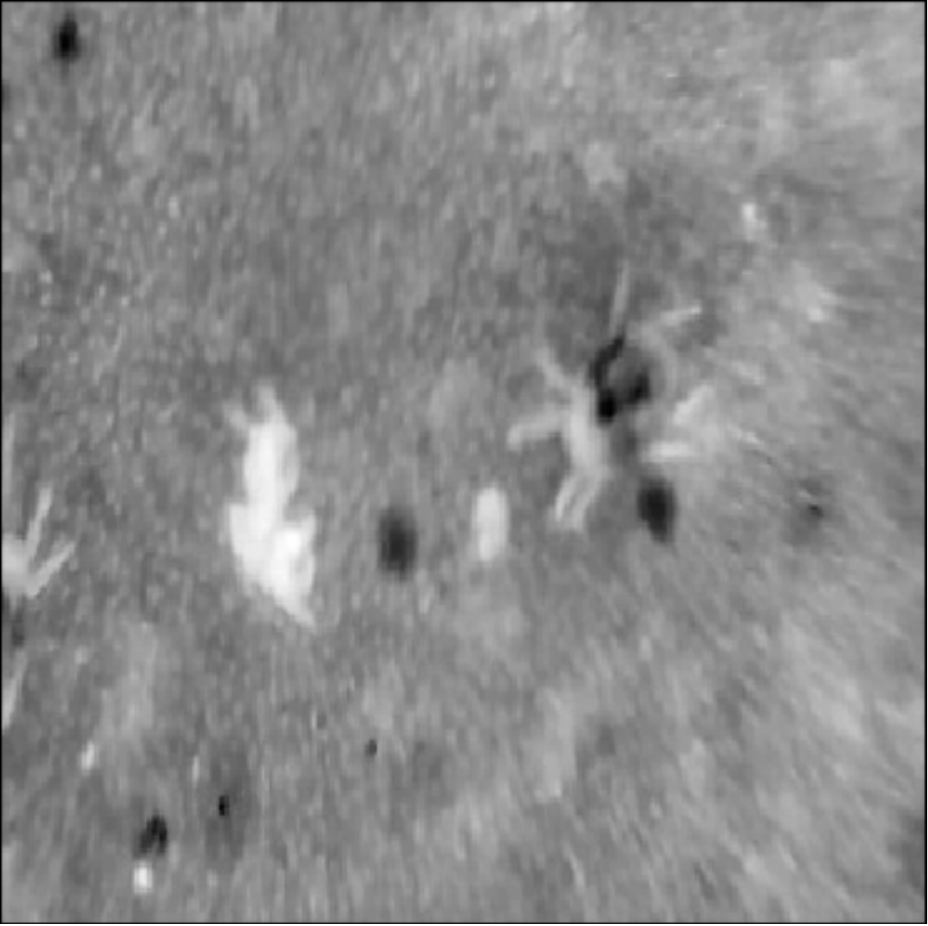}\label{pics:RGB-Transf-b}}
\caption{RGB image transformation. (a) RGB encoded image, $[R,G,B] \in \{0,..., 256\}^{w \times h \times 3}$. (b) Equation~\ref{eq:poly} applied in (a) represented in grayscale, $T([R,G,B]) \in [0,1]^{w \times h}$.}
\label{pics:transf_RGB}
\end{figure}

We can understand the Two-WAM transformation as an integer fusion for a red-green-blue (RGB) image that employs three linear coefficients equals to 1, three exponential coefficients equal to 0, 1, and 2, and the constant value equals to 256. Its result generates a new image illustrated in Figure~\ref{pics:transf_RGB}, where three channels are encoded into a single one (Equation~\ref{eq:poly}). In the equation, the divisor holds the maximum value assumed in the dividend to ensure that $Im(T) \subset [0,1]^{w \times h}$.    
\begin{equation}
\label{eq:poly}
T([R,G,B])= \frac{R \cdot 256^{0} + G \cdot 256^{1}+ B \cdot 256^{2}}{255 \cdot ( 1 + 256 + 256^2)}.
\end{equation}

Two-WAM aims to represent two or more feature spaces in only one, as the integer transformation does. If we use $c=10$, the transformation will learn decimal places. For example, ignoring the denominator, if we take two floating-point values as feature maps ($k = 2$) and we use $f_{1} = 0.25$ and $f_{2} = 0.01$, and we find $\alpha_1=1$, $\alpha_2=1$, $\beta_1=2$, and $  \beta_2=0$, Equation~\ref{eq:trans} yields the value $T_{act} = 0.25 \cdot 10^2 + 0.01 \cdot 10^1 = 25.01$, which shows the two real values in just one. The stochastic gradient descent algorithm~\citep{robbins1951stochastic,kiefer1952stochastic} calculates the best way to transform $k$ float values into just one using $c = 10$.

\section{Experiments and Results}
\label{sec:experiments-results}

This section reports the experiments conducted with our Attention-based MIL-Guided, MIL-Guided~\citep{bollis2020weakly}, WILDCAT~\citep{durand2017wildcat}, and Attention-based Deep MIL~\citep{ilse2018attention}. Subsection~\ref{sec:setup} describes the experimental setup. Subsequently, Subsection~\ref{sec:abalation_study} shows an ablation study to compare the results obtained with Two-WAM (Attention-based MIL-Guided) and Grad-CAM~\citep{selvaraju2017grad} (MIL-Guided). Subsection~\ref{sec:comparison-CPB} compares Attention-based MIL-Guided, WILDCAT, MIL-Guided, and Attention-based Deep MIL considering CPB benchmark. Subsection~\ref{sec:comparison-IP102} presents an analysis of the same weakly supervised learners but using the IP102 dataset. Finally, Subsection~\ref{sec:discussion} discusses our results, providing a quantitative and qualitative analysis.

\subsection{Experimental Setup}
\label{sec:setup}

We use the EfficientNet-B0~\citep{tan2019efficientnet} pre-trained on ImageNet as the main backbone architecture. For Attention-based Deep MIL and WILDCAT approaches, we also consider their original backbones, LeNet~\citep{lecun1998gradient} and ResNet101~\citep{Wu2017a}, respectively.

Following MIL-Guided training~\citep{bollis2020weakly}, we apply the same setup for all experiments using Attention-based MIL-Guided, including five instances in the Instance Model training ($m = 5$). For the Two-WAM mathematical formula, we set $c=10$. For Attention-based Deep MIL\footnote{\url{https://github.com/AMLab-Amsterdam/AttentionDeepMIL}}, we use learning rates range from $10^{-6}$ to $10^{-8}$ and gated attention mechanism (see Subsection~\ref{sec:wsl-rel})~\citep{ilse2018attention} for CPB and IP102 experiments. For WILDCAT\footnote{\url{https://github.com/durandtibo/wildcat.pytorch}}, we consider a learning rate of $2.0\times10^{-3}$, $0.4$ regions, and $8$ maps per class. 

We conducted all experiments on one GPU Nvidia Quadro RTX 8000. The source code used in this work, in addition to detailed descriptions about the data is available in our repository\footnote{\url{https://github.com/edsonbollis}}.

\subsection{Ablation Study on Citrus Pest Benchmark}
\label{sec:abalation_study}
 
We perform an ablation study to analyze the impact of different strategies of our Attention-based MIL-Guided. In this first study, we opted for using only the training and validation set in the analysis of the best strategy to build our models, in order to \textit{not} optimize our hyperparameters on the test set.\vspace{0.1cm}

\noindent \textit{Bag Models' Evaluation:}
Table~\ref{table:bag_model_evaluation} shows the results for the Bag Model, in which we investigated the influence of (1) the removal of noisy images for training (referred to as `NCPB') and (2) dropout on the fully connected layer (referred to as `Drop.'). We set the dropout rate to 30\%. We compare the results of the Attention-based MIL-Guided with Two-WAM (referred to as `Atten.') and MIL-Guided Bag Model~\citep{bollis2020weakly} both on the CPB and the noiseless NCPB validation sets.

The overall picture shows that the `Atten.' approach improves the classification accuracy over the baseline \citep{bollis2020weakly}. Consequently, that illustrates the relevance of the attention-based scheme introduced in this work. Also, comparing the results evaluated on CPB and~NCPB validation sets, they seem to have similar behavior. 

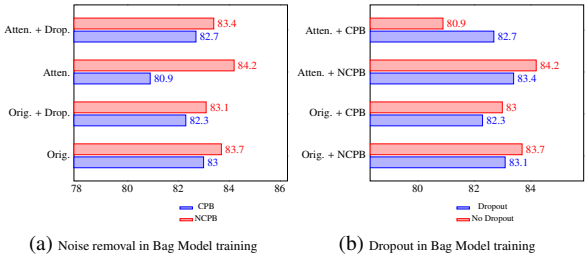
\begin{figure}[t]
\centering
\captionsetup[subfloat]{farskip=2pt,captionskip=2pt}
\hspace*{-0.5cm}\subfloat[\tiny Noise removal in Bag Model training \label{graph:blurred_image_effect}]{
\begin{tikzpicture} [scale=0.41, transform shape]
\begin{axis}[ 
xbar, xmin=80,
xlabel={},
symbolic y coords={%
    Orig., 
    {Orig. + Drop.}, 
    Atten.,     
    {Atten. + Drop.}},
tickwidth         = 0pt,
enlarge y limits  = 0.2,
enlarge x limits  = 0.5,
ytick=data,
nodes near coords, 
nodes near coords align={horizontal},
ytick=data,
legend image code/.code={
        \draw [/tikz/.cd,bar width=2pt,yshift=-1.em,bar shift=1pt]
        plot coordinates {(0.4cm,0.8em)};
    },
    legend style={draw=none, fill=none, font=\footnotesize, at={(0.7,-0.1)}},
    legend cell align=center,
]
\addplot coordinates {
    (83.0,Orig.) 
    (82.3,{Orig. + Drop.})
    (80.9,Atten.)
    (82.7,{Atten. + Drop.})
    };

\addplot coordinates {
    (83.7,Orig.) 
    (83.1,{Orig. + Drop.})
    (84.2,Atten.)
    (83.4,{Atten. + Drop.})
    };

\legend{CPB, NCPB}

\end{axis}
\end{tikzpicture} 
} \hspace*{-0.2cm}
\subfloat[\tiny Dropout in Bag Model training \label{graph:dropout_image_effect_bag_model}]{
\begin{tikzpicture} [scale=0.41, transform shape]
\begin{axis}[ 
xbar, xmin=80,
xlabel={},
symbolic y coords={%
    {Orig. + NCPB}, 
    {Orig. + CPB}, 
    {Atten. + NCPB}, 
    {Atten. + CPB}},
tickwidth         = 0pt,
enlarge y limits  = 0.2,
enlarge x limits  = 0.4,
ytick=data,
nodes near coords, 
nodes near coords align={horizontal},
ytick=data,
legend image code/.code={
        \draw [/tikz/.cd,bar width=2pt,yshift=-1.em,bar shift=1pt]
        plot coordinates {(0.4cm,0.8em)};
    },
    legend style={draw=none, fill=none, font=\footnotesize, at={(0.7,-0.1)}},
    legend cell align=center,
]

\addplot coordinates {
    (82.3,{Orig. + CPB}) 
    (83.1,{Orig. + NCPB})
    (82.7,{Atten. + CPB})
    (83.4,{Atten. + NCPB})
    };

\addplot coordinates {
    (83.0,{Orig. + CPB}) 
    (83.7,{Orig. + NCPB})
    (80.9,{Atten. + CPB})
    (84.2,{Atten. + NCPB})
    };

\legend{Dropout, No Dropout}

\end{axis}
\end{tikzpicture} 
}  

\caption{The effects of using: (a) removal of noisy images and (b) dropout in Bag Model training. Acronyms: `Atten.' models trained using the attention-based activation map proposed approach (Two-WAM); `Drop.' models trained with dropout; `NCPB' experiments trained with only images or instances from NCPB; and `CPB' experiments trained with original images.}
\label{graphs:instance-model}
\end{figure}

Figure~\ref{graph:blurred_image_effect}, which summarizes the results of the `NCPB Validation Set -- Acc' column presented in Table~\ref{table:bag_model_evaluation}, reports better performance using models trained on `NCPB' than ones trained on CPB (all images). We achieved the best result for Attention-based MIL-Guided trained on `NCPB', with 82.3\% accuracy and 79.4\% F1-score on CPB validation, and 84.2\% accuracy and 81.8\% F1-score on NCPB validation.  

In contrast, Figure~\ref{graph:dropout_image_effect_bag_model} shows that dropout (`Drop.') negatively influences Bag Model training in most experiments. Indeed, we observed that dropout alone improves the classification performance, but the combination of `NCPB' and `Drop.' does not outperform both individual strategies. CPB's best dropout results in Table~\ref{table:bag_model_evaluation} are 81.7\% accuracy and 78.6\% F1-score for Attention-based MIL-Guided, and 83.4\% accuracy and 82.0\% F1-score on NCPB (from different configurations).  

Hence, we conclude that the most suitable Bag Model configuration for generating saliency maps is the Attention-based MIL-Guided model trained on NCPB without dropout. Table~\ref{table:bag_model_evaluation} shows that our best configuration improved the baseline by 1 percentage point, achieving 82.3\% accuracy and 79.4\% F1-score. Regarding NCPB validation set, we reached 84.2\% accuracy and 81.8\% F1-score.\vspace{0.1cm}

\noindent \textit{Instance Models' Evaluation:}
Table~\ref{table:instance_model_blurred} shows the results for the Instance Model, in which we conducted experiments to understand the impact of using Bag Model fine-tuning (referred to as `FT') and the removal of noisy images (referred to as `NCPB') on training instances (400$\times$400 pixels). These instances, CPB and NCPB patches, contain relatively larger ROIs compared to the entire images, i.e., more salient mites. Table~\ref{table:instance_model_blurred} presents the results obtained in the CPB and NCPB validation sets, split into two groups according to the activation map method that offers locations for Patch-SaliMap (Subsection~\ref{sec:MIL-Guided}), Grad-CAM and Two-WAM. As in the Bag Models' experiments, we compare Attention-based MIL-Guided with Two-WAM (referred to as `Atten.') with MIL-Guided Bag Model~\citep{bollis2020weakly}. It is worth mentioning that some fine-tuning options are not possible due to the architectural differences, such as fine-tuning an Attention-based MIL-Guided Bag Model like a MIL-Guided Instance Model (referred to as 'Two-WAM' + 'FT' +~'\citep{bollis2020weakly}'). 

\begin{table}[t]
\setlength{\tabcolsep}{1.0mm}
\begin{center}
\caption{Classification accuracy (Acc. in \%) and F1-score (F1 in \%) results for different strategies on CPB and NCPB validation sets. `Atten.' refers to models trained using the attention-based activation map proposed approach (Two-WAM); and `Drop.' models trained with dropout on the fully connected layer. The highlights in bold correspond to the best results in the validation set.} \label{table:bag_model_evaluation}\vspace{0.1cm} 
\footnotesize
\begin{tabular}{ccccccc}
\toprule
& \multirow{2}{*}{NCPB} & \multirow{2}{*}{Drop.} & \multicolumn{2}{c}{CPB Validation Set} & \multicolumn{2}{c}{NCPB Validation Set} \\
& & & Acc. (\%) & F1 (\%) & Acc. (\%) & F1 (\%) \\
\midrule
\citep{bollis2020weakly} & & &  $80.9$ \tiny{$\pm 1.9$} &  $78.4$ \tiny{$\pm 2.1$} & $83.0$ \tiny{$\pm 0.9$} & $80.8$ \tiny{$\pm 0.8$} \\
\citep{bollis2020weakly} & & $\bullet$ &  $81.2$ \tiny{$\pm 1.1$}  &  $78.4$ \tiny{$\pm 1.3$} & $82.3$ \tiny{$\pm 0.9$} & $\textbf{82.0}$ \tiny{$\pm \textbf{0.8}$} \\
\citep{bollis2020weakly} & $\bullet$ &  &  $81.4$ \tiny{$\pm 0.9$} &  $79.2$ \tiny{$\pm 1.1$} & $83.7$ \tiny{$\pm 0.9$} & $81.8$ \tiny{$\pm 0.8$} \\
\citep{bollis2020weakly} & $\bullet$ & $\bullet$ &  $80.7$ \tiny{$\pm 1.6$} &  $78.6$ \tiny{$\pm 1.7$} & $83.1$ \tiny{$\pm 1.8$} & $81.0$ \tiny{$\pm 1.9$} \\
\cmidrule(lr){1-7}
Atten. &  &  &  $80.7$ \tiny{$\pm 1.3$} &  $76.8$ \tiny{$\pm 2.3$} & $80.9$ \tiny{$\pm 1.1$} & $77.2$ \tiny{$\pm 1.3$} \\
Atten. &  & $\bullet$ &  $81.7$ \tiny{$\pm 1.3$} &  $77.8$ \tiny{$\pm 2.0$} & $82.7$ \tiny{$\pm 1.1$} & $79.2$ \tiny{$\pm 1.9$} \\
Atten. & $\bullet$ &  &  $\textbf{82.3}$ \tiny{$\pm \textbf{1.5}$} &  $\textbf{79.4}$ \tiny{$\pm 1.8$} & $\textbf{84.2}$ \tiny{$\pm \textbf{1.4}$} & $81.8$ \tiny{$\pm 1.3$} \\
Atten. & $\bullet$ & $\bullet$ & $81.5$ \tiny{$\pm 1.2$} &  $78.6$ \tiny{$\pm 1.1$} & $83.4$ \tiny{$\pm \textbf{1.4}$} & $80.4$ \tiny{$\pm 1.5$} \\
\bottomrule
\end{tabular}
\end{center}
\end{table}

Most Two-WAM-based configurations improve the classification performance over the Grad-CAM baseline. Once again, that result illustrates the relevance of the attention-based layer introduced in this work. The best result reached with CPB instances produced by the Grad-CAM were 91.8\% accuracy and 91.0\% F1-score, our previous work~\citep{bollis2020weakly}. 

\begin{figure}[thb]
\centering
\captionsetup[subfloat]{farskip=2pt,captionskip=2pt}
 \hspace*{-0.5cm}\subfloat[\tiny Noise in Instance Model training \label{graph:noise_instance_model_effect}]{
    \begin{tikzpicture} [scale=0.41, transform shape]
    \begin{axis}[ 
    xbar, xmin=80,
    xlabel={},
    symbolic y coords={%
        Orig., 
        {Orig.*}, 
        Atten., 
        {Atten. + FT}},
    tickwidth         = 0pt,
    enlarge y limits  = 0.2,
    enlarge x limits  = 0.4,
    ytick=data,
    nodes near coords, 
    nodes near coords align={horizontal},
    ytick=data,
    legend image code/.code={
            \draw [/tikz/.cd,bar width=2pt,yshift=-1.em,bar shift=1pt]
            plot coordinates {(0.4cm,0.8em)};
        },
        legend style={draw=none, fill=none, font=\footnotesize, at={(0.7,-0.1)}},
        legend cell align=center,
    ]
    \addplot coordinates {
        (93.5,Orig.) 
        (87.9,{Orig.*})
        (94.2,Atten.)
        (92.9,{Atten. + FT})
        };
    
    \addplot coordinates {
        (91.2,Orig.) 
        (85.6,{Orig.*})
        (92.2,Atten.)
        (89.7,{Atten. + FT})
        };

    \legend{CPB, NCPB}
    
    \end{axis}
    \end{tikzpicture}
\label{graph:dropout_image_effect_instance_model}
} \hspace*{-0.2cm}
\subfloat[\tiny Bag Model fine-tuned in the Instance Model training\label{graph:ft_instance_model}]{
\begin{tikzpicture} [scale=0.41, transform shape]
    \begin{axis}[ 
    xbar, xmin=80,
    xlabel={},
    symbolic y coords={%
    {Atten. + CPB*},
    {Atten. + NCPB*},
    {Atten. + CPB},
    {Atten. + NCPB},},
    tickwidth         = 0pt,
    enlarge y limits  = 0.2,
    enlarge x limits  = 0.4,
    ytick=data,
    nodes near coords, 
    nodes near coords align={horizontal},
    ytick=data,
    legend image code/.code={
        \draw [/tikz/.cd,bar width=2pt,yshift=-1.em,bar shift=1pt]
        plot coordinates {(0.4cm,0.8em)};
    },
    legend style={draw=none, fill=none, font=\footnotesize, at={(0.7,-0.1)}},
    legend cell align=center,
    ]

    \addplot coordinates {
    (89.6,{Atten. + CPB*})
    (87.5,{Atten. + NCPB*})
    (92.9,{Atten. + CPB})
    (89.7,{Atten. + NCPB})
    };
    
    \addplot coordinates {
    (88.8,{Atten. + CPB*})
    (85.5,{Atten. + NCPB*})
    (94.2,{Atten. + CPB})
    (92.2,{Atten. + NCPB})
    };
    
    \legend{Fine-tuned, No Fine-tuned}
    
    \end{axis}
    \end{tikzpicture} 
}
\caption{The effects of using: (a) removal of noisy images and (b) Bag Model fine-tuned in the Instance Model training. `*' stands for the experiments using Grad-CAM to produce instances for Instance Models while no mark indicates the use of Two-WAM. `Atten.' refers to models trained using the attention-based activation map proposed approach (Two-WAM); `NCPB' experiments trained with only images or instances from NCPB; `CPB' experiments trained on all images or its instances; and `FT' the Bag Model fine-tuned experiments in the Instance Model training.}
\label{graphs:bag-model}
\end{figure}
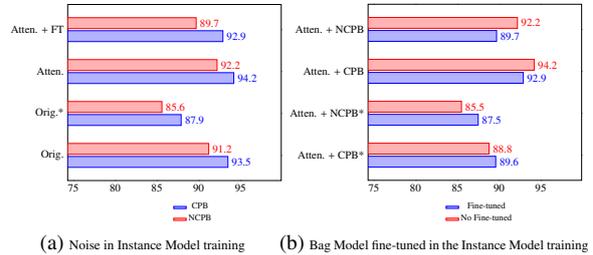

Figure~\ref{graph:noise_instance_model_effect} shows the best performance on models that consider CPB or NCPB images for training (`NCPB Validation Set -- Acc' column). This behavior is the opposite obtained in the Bag Models' experiments. Thus, we must carefully handle tiny ROI in noise presence from images captured under natural conditions. In addition, our results suggested that salient ROI would benefit itself from noise. The best result for Attention-based MIL-Guided trained on `NCPB' instances are 91.7\% accuracy and 90.6\% F1-score on CPB, and 92.2\% accuracy and 91.2\% F1-score on NCPB.

Figure~\ref{graph:ft_instance_model} presents the results concerning the Bag Model fine-tuning in the Instance Model training. The fine-tuning does not improve the classification performance using Two-WAM instances, but it improves using Grad-CAM instances. The fine-tuning strategy's best result is 92.8\% accuracy and 92.2\% F1-score in CPB validation set, and 92.9\% accuracy and 91.8\% F1-score on the NCPB validation set. We achieved the best result for Instance Model using Two-WAM and all CPB images to train models. It improves the accuracy by 2.2~percentage points, reaching 94.0\% accuracy and 93.4\% F1-score in the CPB validation set. In the following experiments, we used that configuration. 

\begin{table}[t]
\setlength{\tabcolsep}{1.0mm}
\begin{center}
\caption{Classification accuracy (Acc. in \%) and F1-score (F1. in \%) results for different strategies in the instance set generated from CPB and NCPB validation sets. `Atten.' refers to models trained using the proposed attention-based activation map approach (Two-WAM); and `FT' the Bag Model fine-tuned experiments in the Instance Model training. The highlights in bold correspond to the best results in the validation set.}
\vspace{0.1cm}
\label{table:instance_model_blurred}
\footnotesize
\begin{tabular}{lccccccc}
\toprule
& & \multirow{2}{*}{NCPB} & \multirow{2}{*}{FT} & \multicolumn{2}{c}{CPB Validation Set} & \multicolumn{2}{c}{NCPB Validation Set} \\
& &     &    & Acc. (\%) & F1 (\%) & Acc. (\%) & F1 (\%)\\
\midrule
\multirow{7}{*}{\rotatebox{90}{Grad-CAM}} & \citep{bollis2020weakly} & & $\bullet$ & $91.8$ \tiny{$\pm 2.4$} & $91.0$ \tiny{$\pm 2.2$} & -- & -- \\
& \citep{bollis2020weakly} & & & $88.0$ \tiny{$\pm 0.8$} & $86.8$ \tiny{$\pm 1.3$}  & $87.9$ \tiny{$\pm 0.6$} & $86.6$ \tiny{$\pm 0.5$}  \\
& \citep{bollis2020weakly} & $\bullet$ & & $85.6$ \tiny{$\pm 1.0$} & $84.4$ \tiny{$\pm 1.1$}  & $85.6$ \tiny{$\pm 1.2$} & $84.4$ \tiny{$\pm 1.1$}  \\

& Atten. & &  & $89.0$ \tiny{$\pm 1.1$} & $87.8$ \tiny{$\pm 1.3$}  & $88.8$ \tiny{$\pm 1.3$} & $87.8$ \tiny{$\pm 1.3$}  \\
& Atten. & & $\bullet$ & $89.3$ \tiny{$\pm 0.6$} & $88.0$ \tiny{$\pm 0.7$}  & $89.6$ \tiny{$\pm 0.7$} & $88.6$ \tiny{$\pm 0.9$}  \\
& Atten. & $\bullet$ & & $86.6$ \tiny{$\pm 2.0$} & $85.2$ \tiny{$\pm 2.2$}  & $85.5$ \tiny{$\pm 2.6$} & $84.4$ \tiny{$\pm 2.9$}  \\
& Atten. & $\bullet$ & $\bullet$ & $86.7$ \tiny{$\pm 3.4$} & $84.2$ \tiny{$\pm 5.0$}  & $87.5$ \tiny{$\pm 3.2$} & $85.8$ \tiny{$\pm 4.1$}  \\ \cmidrule(lr){1-8}
\multirow{6}{*}{\rotatebox{90}{Two-WAM}} & \citep{bollis2020weakly} & & & $93.3$ \tiny{$\pm 0.8$} & $92.2$ \tiny{$\pm 0.8$}  & $93.5$ \tiny{$\pm 0.7$} & $92.6$ \tiny{$\pm 0.5$}  \\
& \citep{bollis2020weakly} & $\bullet$ & & $90.4$ \tiny{$\pm 0.9$} & $89.2$ \tiny{$\pm 0.8$}  & $91.2$ \tiny{$\pm 1.0$} & $90.4$ \tiny{$\pm 1.1$} \\
& Atten. &  & & $\textbf{94.0}$ \tiny{$\pm \textbf{0.6}$} & $\textbf{93.4}$ \tiny{$\pm \textbf{0.5}$}  & $\textbf{94.2}$ \tiny{$\pm \textbf{0.6}$} & $\textbf{93.2}$ \tiny{$\pm \textbf{0.4}$}  \\
& Atten. &  & $\bullet$ & $92.8$ \tiny{$\pm 0.6$} & $92.2$ \tiny{$\pm 0.4$}  & $92.9$ \tiny{$\pm 0.6$} & $91.8$ \tiny{$\pm 0.8$}  \\
& Atten.	& $\bullet$ &  & $91.7$ \tiny{$\pm 1.1$} & $90.6$ \tiny{$\pm 0.9$}  & $92.2$ \tiny{$\pm 0.8$} & $91.2$ \tiny{$\pm 1.1$}  \\
& Atten.	& $\bullet$ & $\bullet$ & $88.3$ \tiny{$\pm 1.2$} & $87.2$ \tiny{$\pm 1.1$}  & $ 89.7$ \tiny{$\pm 2.7$} & $88.6$ \tiny{$\pm 2.9$}  \\
\bottomrule
\end{tabular}
\end{center}
\end{table}

\subsection{Weakly Supervised Methods Applied to Citrus Pest Benchmark}
\label{sec:comparison-CPB}

We compare Attention-based MIL-Guided to WSL state-of-the-art methods, Attention-based Deep MIL and WILDCAT, using the CPB test set. We explore two scenarios: (1) CPB image sizes of 800$\times$800 pixels, no zoom augmentation, and batch size equals to the number of instances of one image, and (2) CPB image sizes of 1,200$\times$1,200 pixels (original size), and the maximum batch size that one GPU supports for each architecture. We choose these two scenarios due to memory overhead in Attention-based Deep MIL training using LeNet as the backbone. The entire architecture contains 552 million parameters (see Table~\ref{table:ReportesCPB}). This fact, along with the image size (1,200$\times$1,200 pixels), made it impracticable to conduct this experiment on the available GPU.

\begin{table*}[t]
\setlength{\tabcolsep}{1.0mm}
\begin{center}
\caption{Classification accuracy (Acc. in \%) and F1-score (F1. in \%) of different WSL on the CPB test set, all models used the NCPB in training. W. (in~M) means the number of weights in millions of each CNN. `Trained on NCPB' means that we trained using NCPB and evaluated on CPB, and `Best' refers to our best training process. The highlights in bold correspond to the best results.}
\vspace{0.1cm}
\label{table:ReportesCPB}
\begin{tabular}{clccccc}
\toprule
&                    & \multicolumn{2}{c}{800$\times$800} & \multicolumn{2}{c}{1200$\times$1200} &  \\
& WSL & Acc. (\%) & F1 (\%) & Acc. (\%) & F1 (\%) & W. (M) \\
\midrule
\multirow{6}{*}{\rotatebox{90}{\footnotesize{Trained on NCPB}}}
& Attention-based Deep MIL (LeNet)~\citep{ilse2018attention} & $63.6 $ \tiny{$\pm 0.5$} & $62.2$ \tiny{$\pm 0.8$} & $-$ & $-$ & $552.0$\\
& Attention-based Deep MIL (EfficientNet) & $63.3$ \tiny{$\pm 4.1$} & $66.8$ \tiny{$\pm 5.4$} & $67.1$ \tiny{$\pm 3.3$} & $63.4$ \tiny{$\pm 2.7$}	 & $10.0$\\
& WILDCAT (ResNet101)~\citep{durand2017wildcat} & $65.5 $ \tiny{$\pm 4.9$} & $55.4$ \tiny{$\pm 2.6$} & $71.9$ \tiny{$\pm 3.3$} & $68.2$ \tiny{$\pm 6.9$}	& $42.5$\\
& WILDCAT (EfficientNet)  & $70.0$ \tiny{$\pm 2.6$} & $67.9$ \tiny{$\pm 1.9$} & $76.4$ \tiny{$\pm 0.2$} & $73.0$ \tiny{$\pm 1.2$} & $\textbf{4.03}$\\
& Attention-based MIL-Guided (Bag Model) & $74.1$ \tiny{$\pm 3.4$} & $72.4$ \tiny{$\pm 3.1$} & $79.2$ \tiny{$\pm 1.5$} & $76.6$ \tiny{$\pm 1.5$} & $4.05$\\
& Attention-based MIL-Guided (Bag + Inst. Models) & $78.1$ \tiny{$\pm 2.3$} &
$76.8$ \tiny{$\pm 1.9$} & $90.2$ \tiny{$\pm 1.0$} & $89.0$ \tiny{$\pm 1.0$} & $8.1$\\ \midrule
\multirow{2}{*}{\rotatebox{90}{\footnotesize{Best}}}
& MIL-Guided (Bag + Inst. Models) \citep{bollis2020weakly} & $78.8$ \tiny{$\pm 1.8$} & $73.4$ \tiny{$\pm 3.3$} & $90.9$ \tiny{$\pm 1.2$} & $89.0$ \tiny{$\pm 1.6$} & $8.1$\\
& Attention-based MIL-Guided (Bag + Inst. Models) & $\textbf{82.1}$ \tiny{$\pm \textbf{1.2}$} &
$\textbf{80.1}$ \tiny{$\pm \textbf{1.1}$} & $\textbf{92.4}$ \tiny{$\pm \textbf{0.7}$} & $\textbf{91.8}$ \tiny{$\pm \textbf{0.8}$} & $8.1$\\ 
\bottomrule
\end{tabular}
\end{center}
\end{table*}

Attention-based Deep MIL requires instances for training models (equal patches dividing the original image without overlap) and all instances represent a bag or batch. For this reason, we cut images into the exact size that we train the Attention-based MIL-Guided Instance Models, that is, (1) 9 instances with the largest patch sizes and (2) over 500 instances with patch sizes of 32$\times$32 pixels, the smallest possible size as suggested by Ilse et al.~\citep{ilse2018attention}. For the sake of comparison, we consider the highest score reached between the two types of cuts. In the first scenario ($800 \times 800$), we used patch sizes of 266$\times$266 pixels, totaling 9 instances, and 32$\times$32 pixels, resulting in 625 instances. In the second scenario ($1200 \times 1200$), we cut images into 9 instances of 400$\times$400 pixels and 1444 instances of 32$\times$32 pixels. For the Attention-based MIL-Guided, we used 5 instances of 266$\times$266 pixels for the first scenario and 5 instances of 400$\times$400 pixels for the second scenario, as MIL-Guided. 

We organized our results into two groups: (1) we train all models on NCPB and evaluate them on CPB test set, and (2) we compare the results of our best model reached in Subsection~\ref{sec:abalation_study} (Bag model trained on the NCPB and Instance Model trained on CPB) with the best result from CPB literature~\citep{bollis2020weakly}. Our Attention-based MIL-Guided (Bag + Instance Models, 92.4\% accuracy in Table~\ref{table:ReportesCPB}) surpasses the Attention-based Deep MIL and WILDCAT in all scenarios up to 25.3 percentage points. Comparing the Attention-based MIL-Guided (Bag Model, 79.2\% accuracy) with state-of-the-art methods, we outperform both Attention-based Deep MIL and WILDCAT up to 12.1 percentage points. 

The best result for Attention-based Deep MIL, in both scenarios, is 67.1\%  accuracy and 63.4\% F1-score, using 9 instances on 1,200$\times$1,200 pixels. The result for small instances (32$\times$32 pixels) is 63.3\% accuracy and 66.8\% F1-score, using 25 instances on 800$\times$800 pixels. For WILDCAT, the best result is 76.4\% accuracy and 73.0\% F1-score. Our Attention-based MIL-Guided (Bag Model) provides better activation maps and results than literature methods, reaching 79.2\% accuracy and 76.6\% F1-score. Considering our overall best result, in which the Bag Model is trained on NCPB images and the Instance Model on CPB images, we achieved 92.4\% accuracy and 91.8\% F1-score, surpassing the result for MIL-Guided, 90.9\% accuracy and 89.0\% F1-score. Furthermore, the difference of up to 12.4 percentage points between the Instance Model (89.0\%) and the Bag Model (76.6\%) shows the classification performance of the Attention-based MIL-Guided process.

It is noteworthy the difference --- at least 5 percentage points --- between the results of the two scenarios, 800$\times$800 pixels and 1,200$\times$1,200 pixels. The size of the images (800$\times$800 pixels) and the non-application of zoom augmentation, which negatively impacts the capacity of networks to learn how to recognize ROI locations, decrease the classification performance.

\begin{figure}[t]
\captionsetup[subfloat]{farskip=2pt,captionskip=2pt}
\centering

\subfloat[CPB samples~\citep{bollis2020weakly}]{
\includegraphics[height=1.75cm,width=1.75cm]{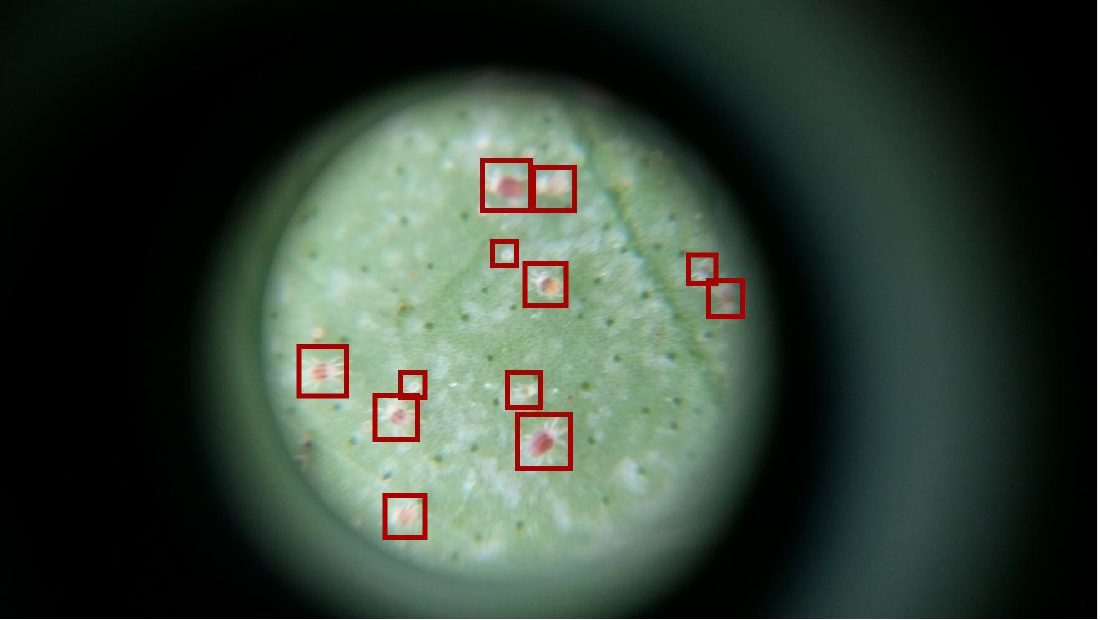} 
\hspace*{0.01cm}
\includegraphics[height=1.75cm,width=1.75cm]{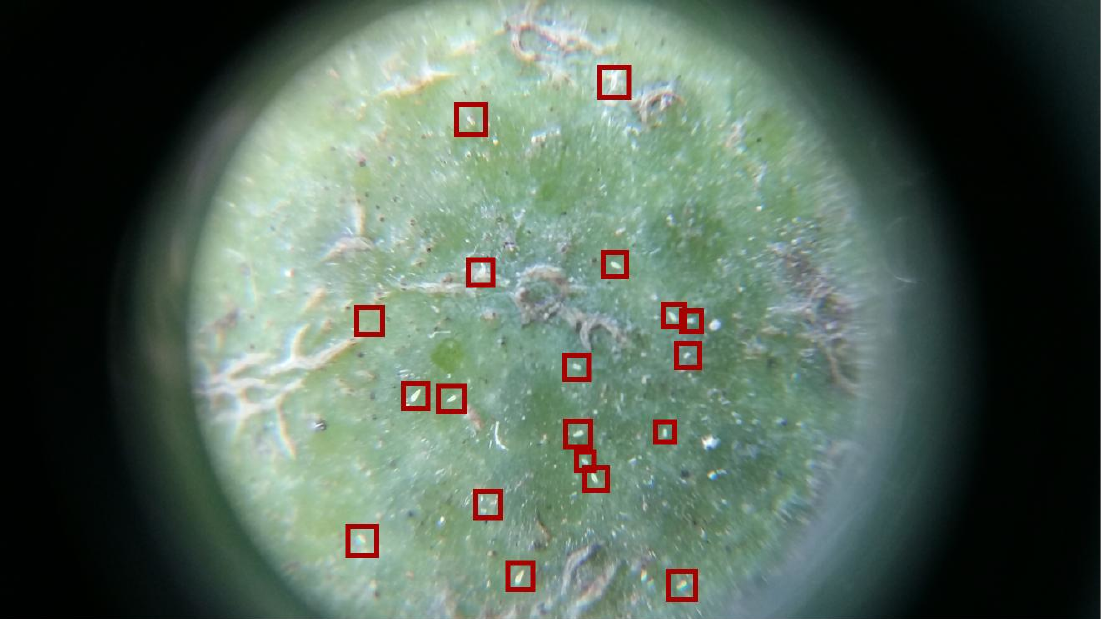}
\hspace*{0.01cm}
\includegraphics[height=1.75cm,width=1.75cm]{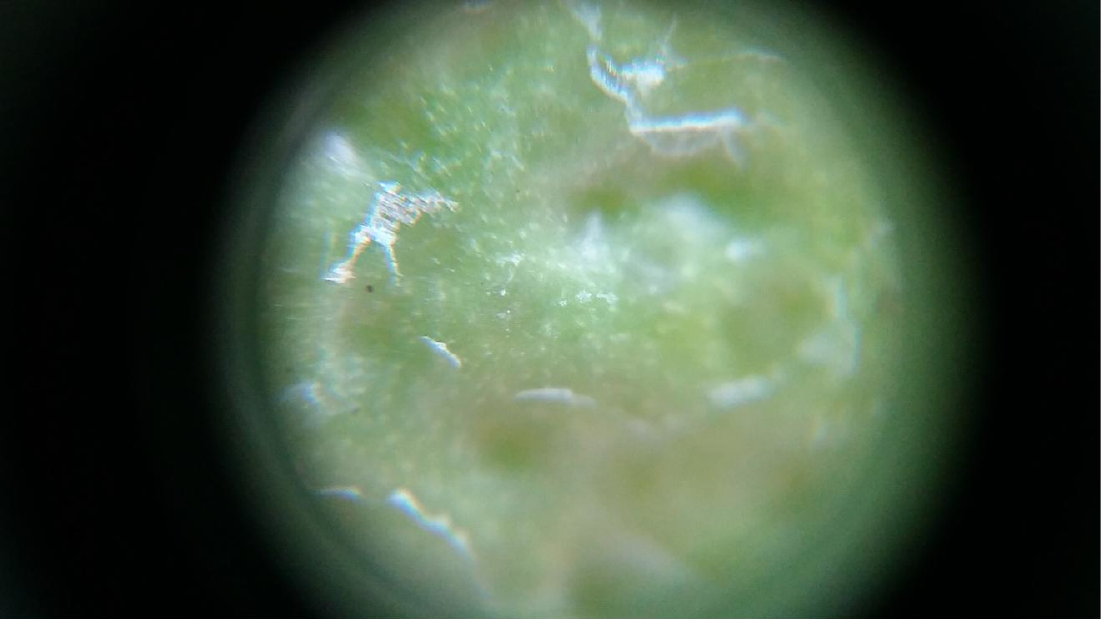}
\hspace*{0.01cm}
\includegraphics[height=1.75cm,width=1.75cm]{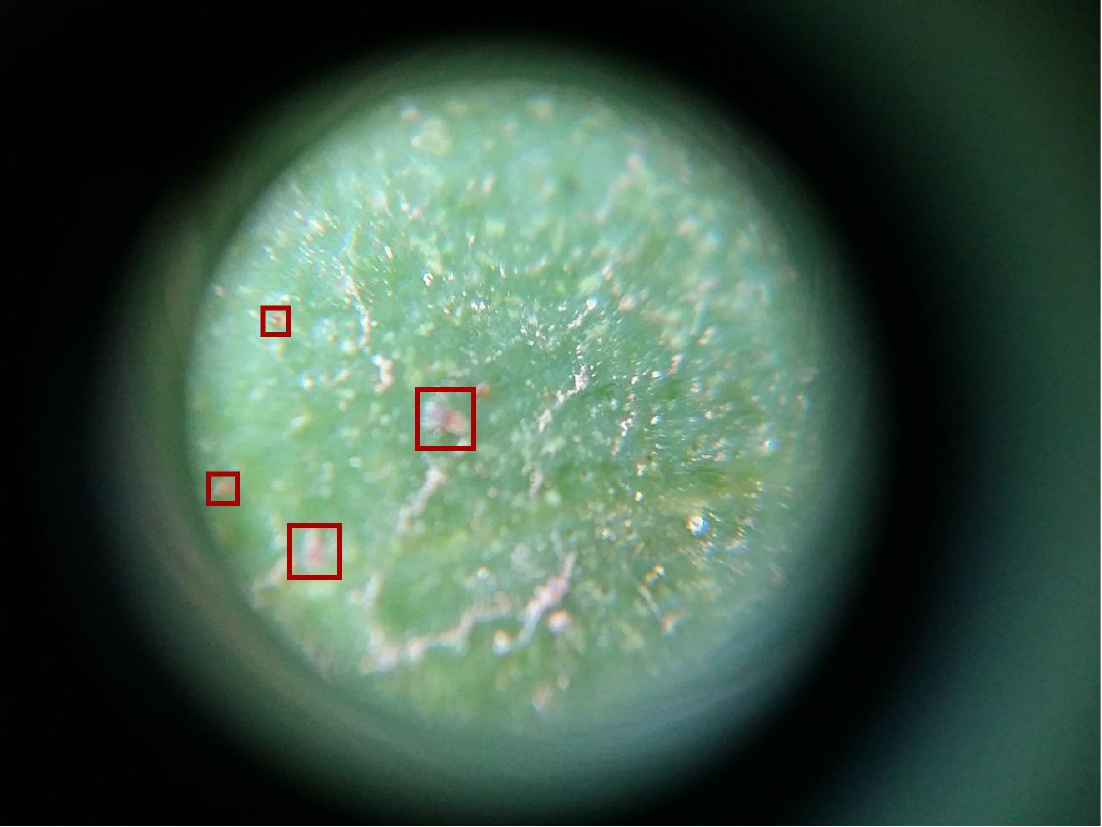}\label{pics:activation-maps-a}}\\

\subfloat[Attention-based MIL-Guided (Two-WAM)]{
\includegraphics[height=1.75cm,width=1.75cm]{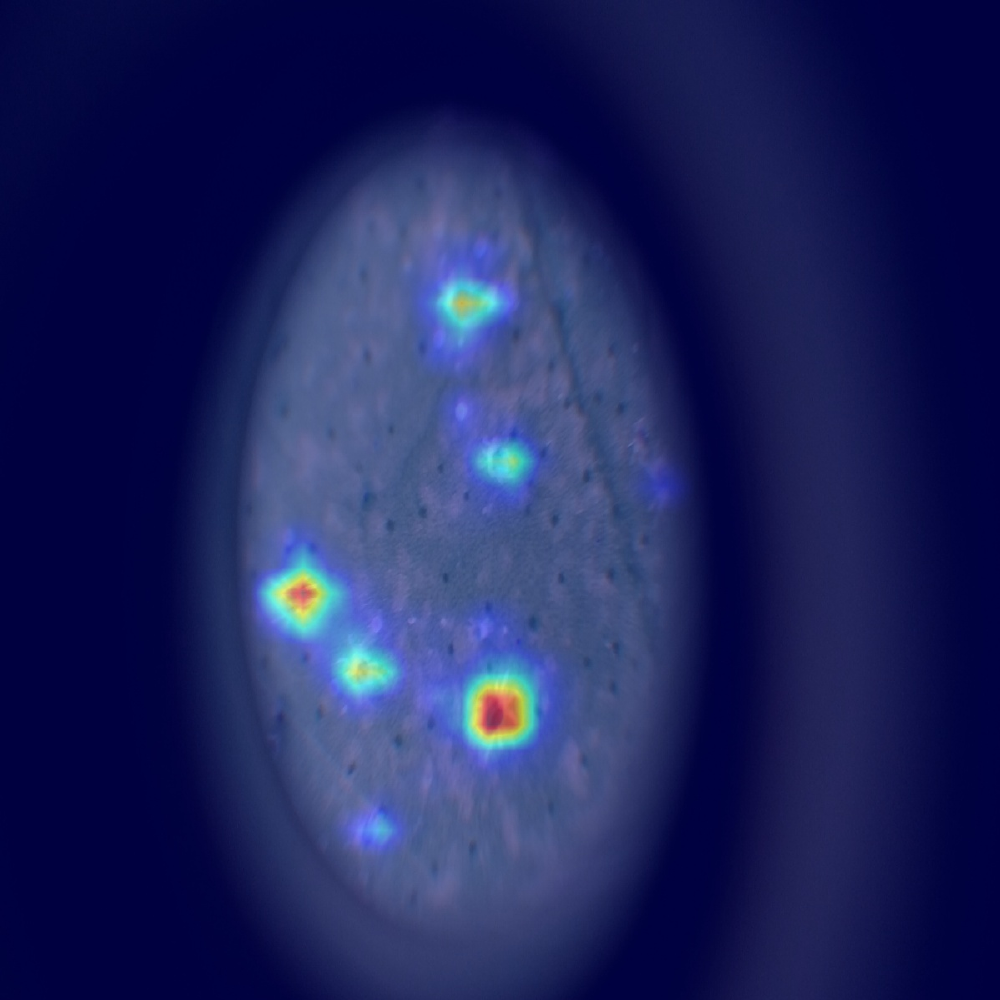}  
\hspace*{0.01cm}
\includegraphics[height=1.75cm,width=1.75cm]{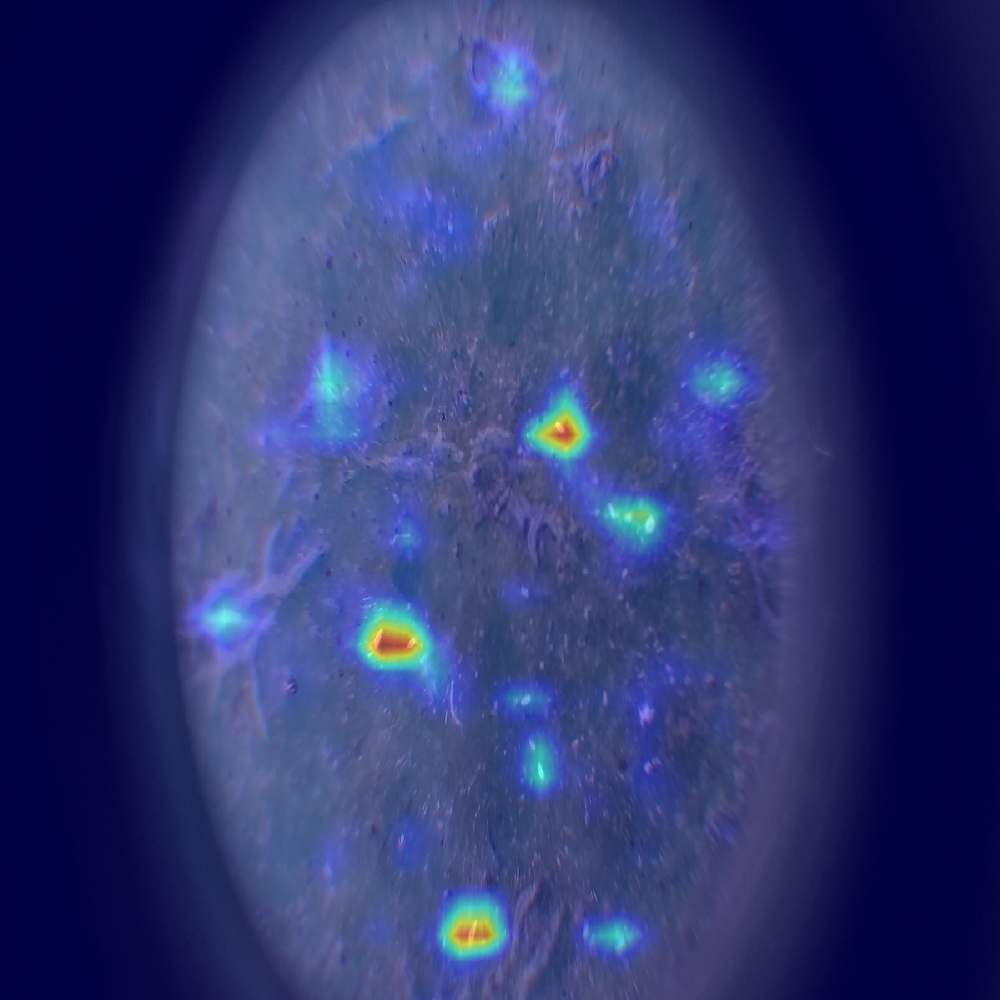}
\hspace*{0.01cm}
\includegraphics[height=1.75cm,width=1.75cm]{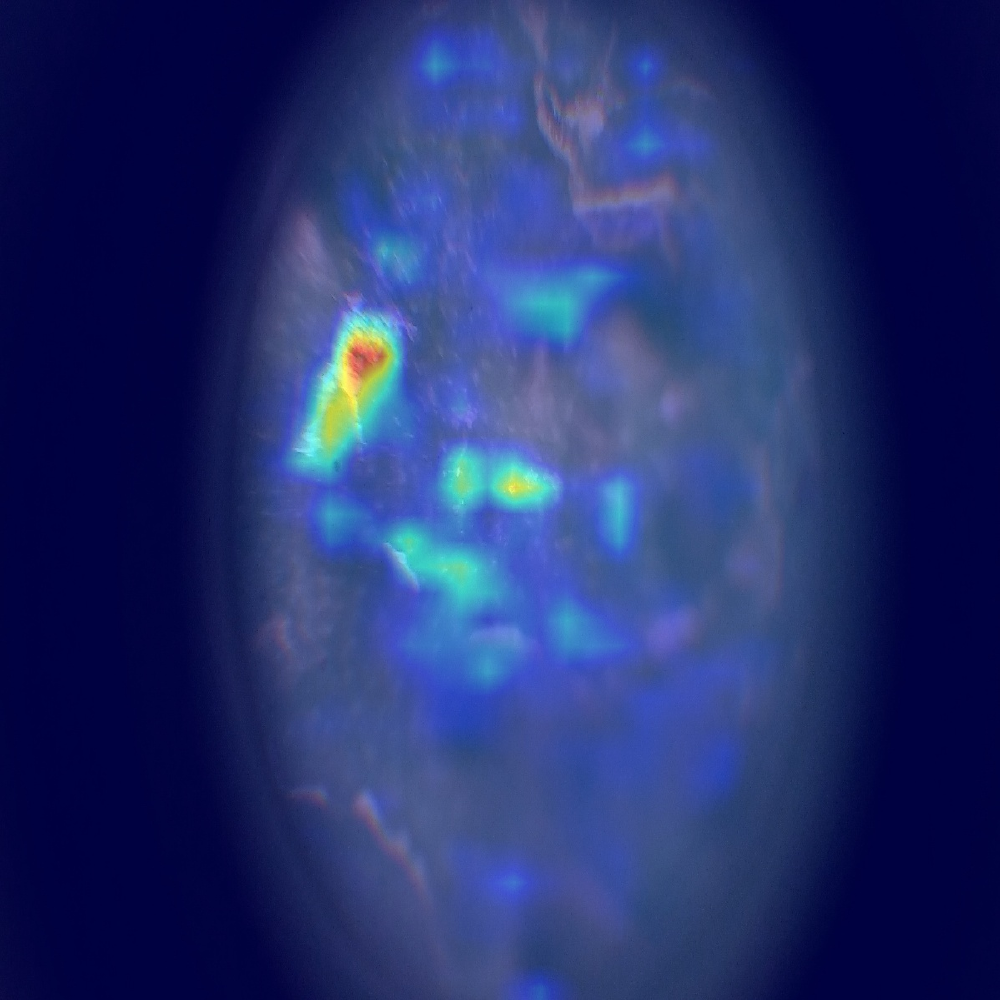}
\hspace*{0.01cm}
\includegraphics[height=1.75cm,width=1.75cm]{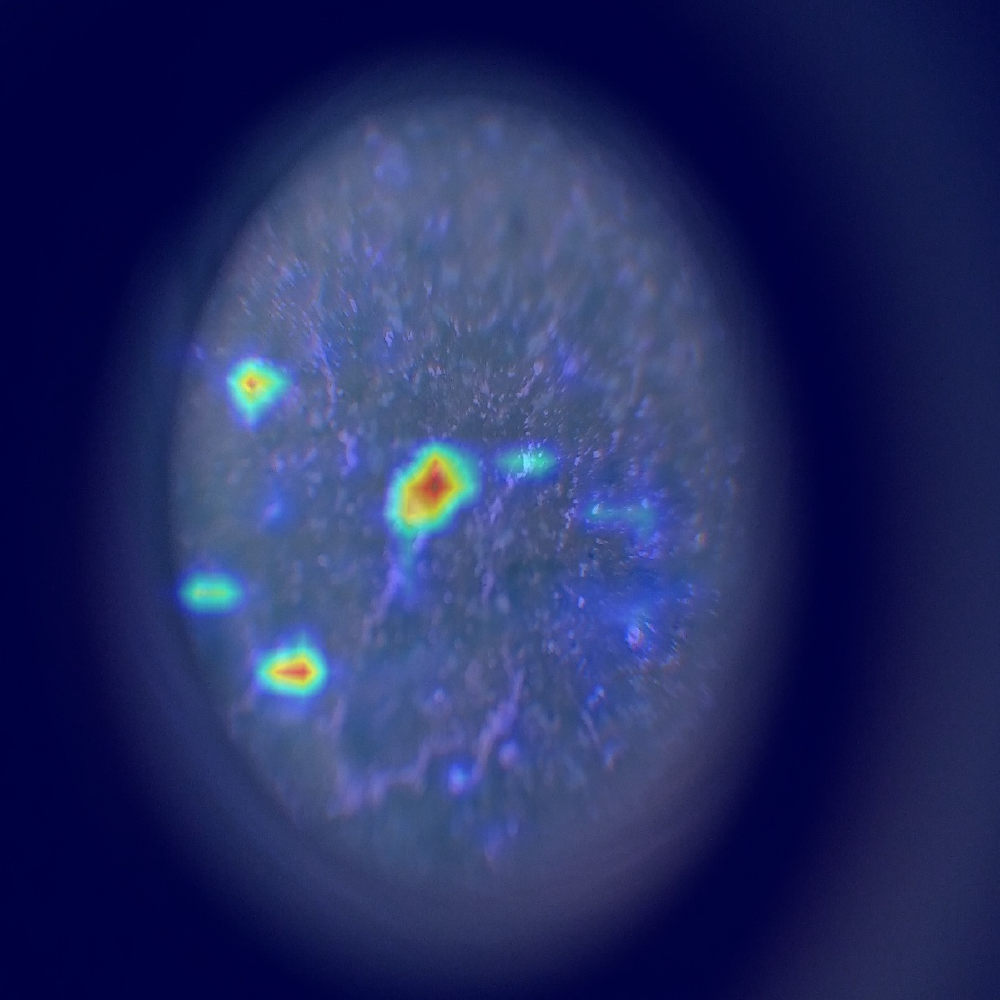}\label{pics:activation-maps-b}}  \\

\subfloat[MIL-Guided (Grad-CAM) \citep{bollis2020weakly}]{
\includegraphics[height=1.75cm,width=1.75cm]{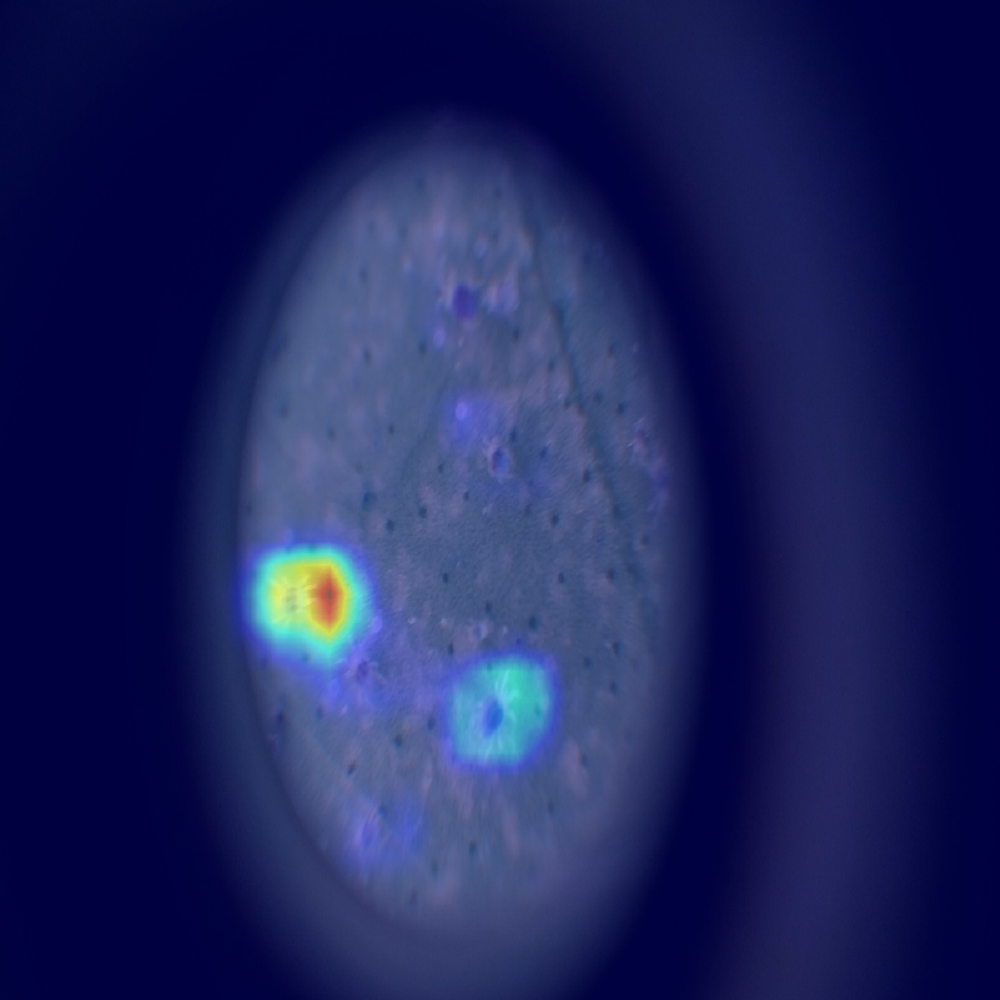}
\hspace*{0.01cm}
\includegraphics[height=1.75cm,width=1.75cm]{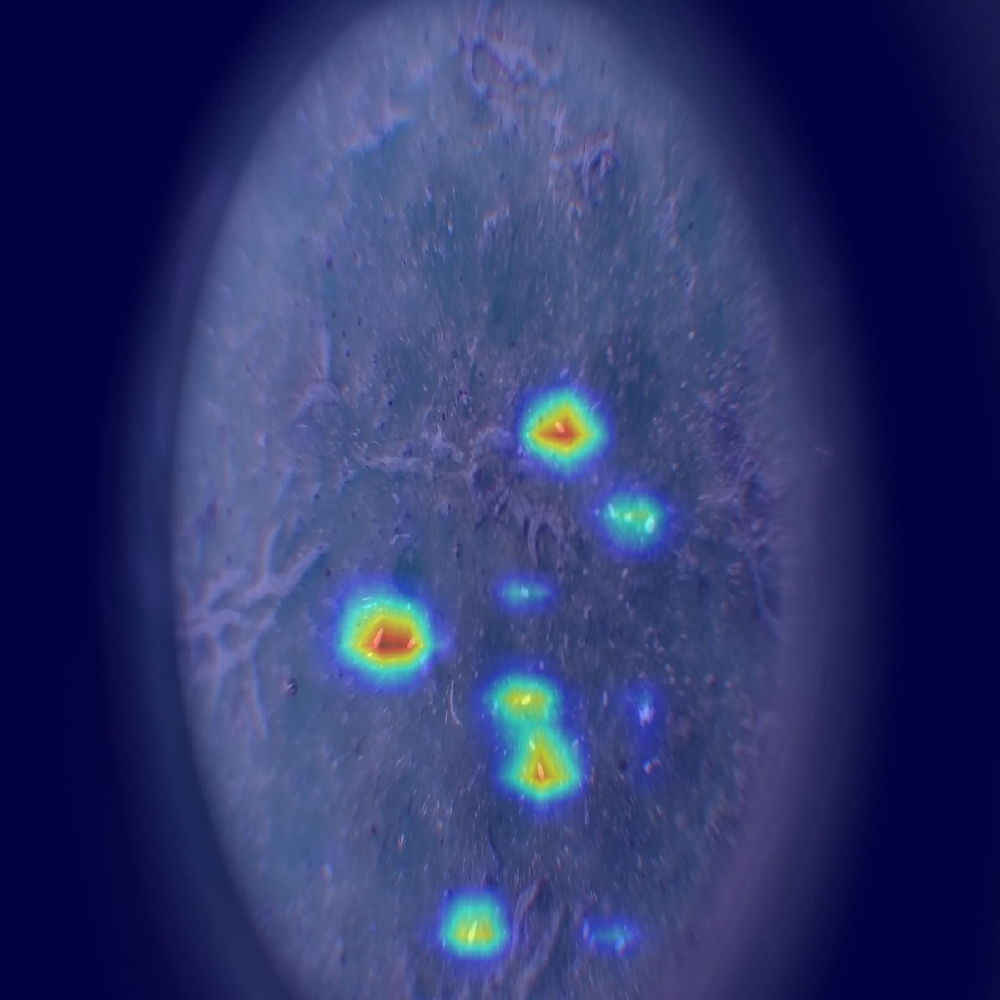} 
\hspace*{0.01cm}
\includegraphics[height=1.75cm,width=1.75cm]{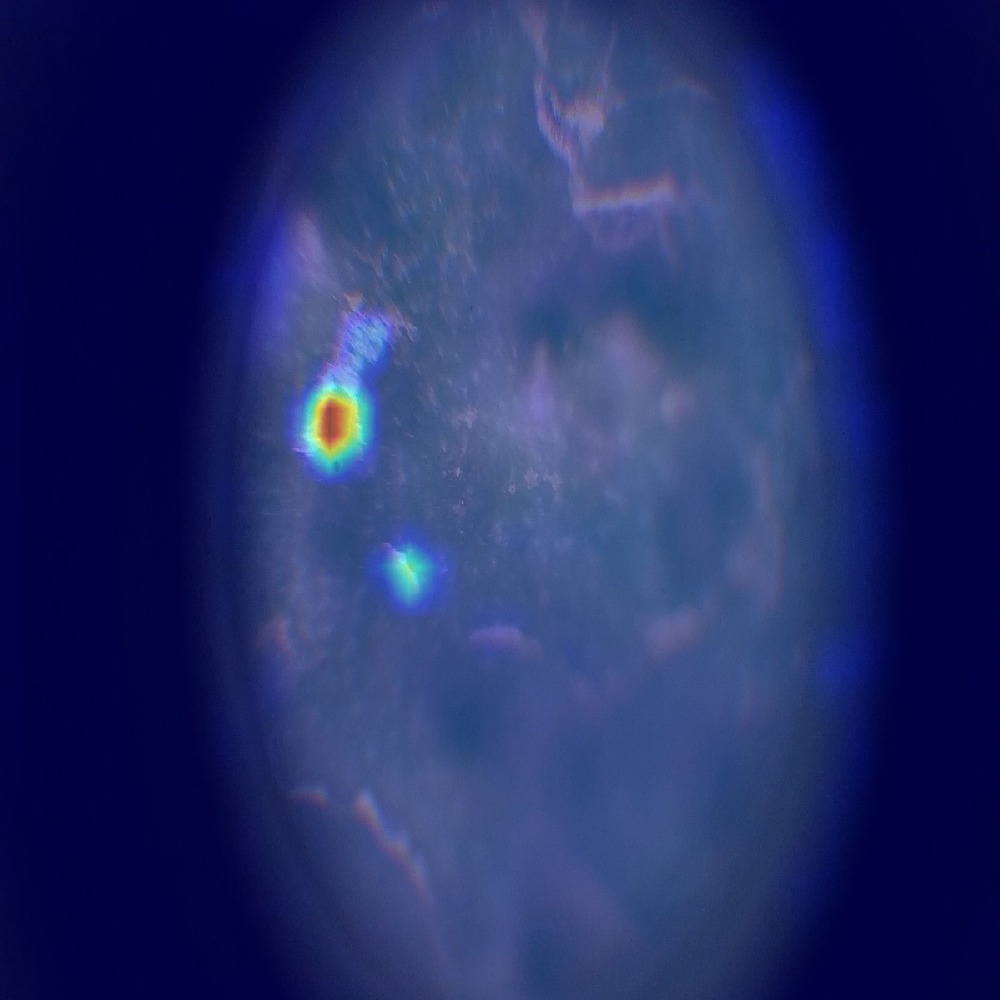}
\hspace*{0.01cm}
\includegraphics[height=1.75cm,width=1.75cm]{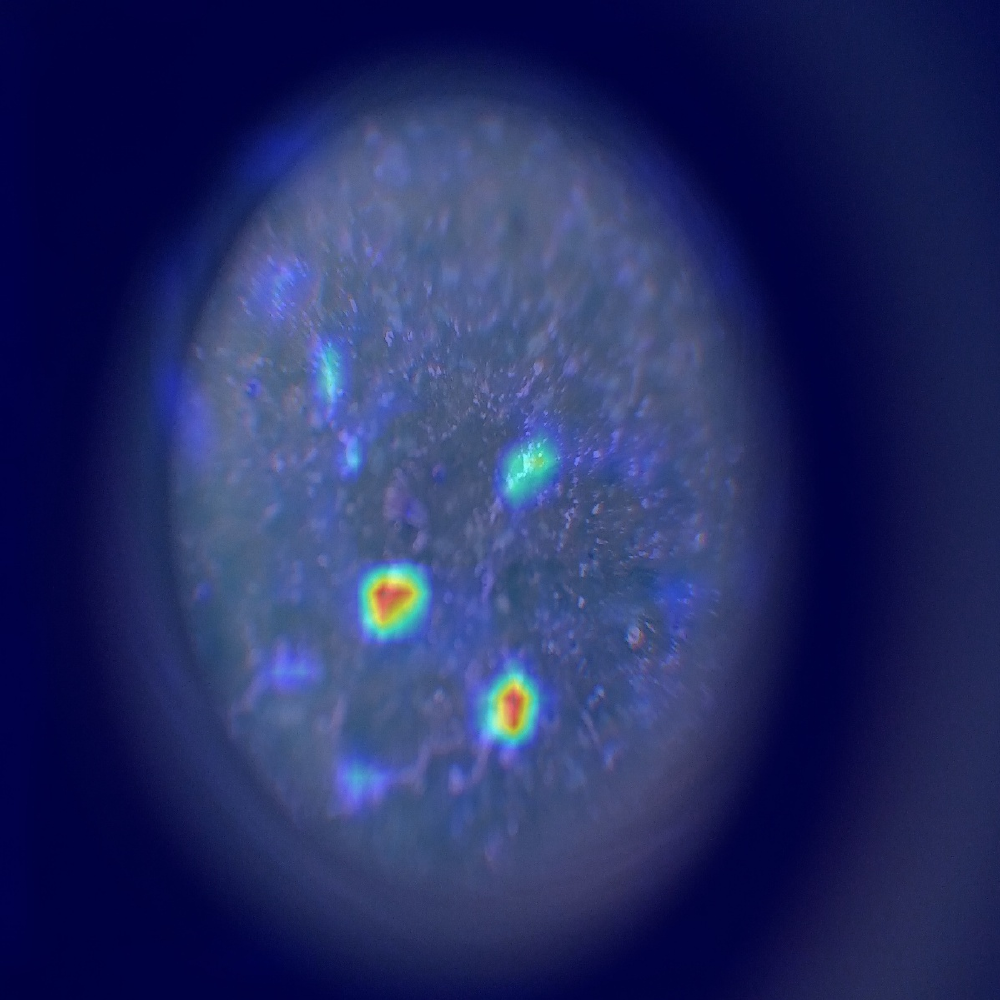} \label{pics:activation-maps-c}} \\

\subfloat[WILDCAT \citep{durand2017wildcat}]{
\includegraphics[height=1.75cm,width=1.75cm]{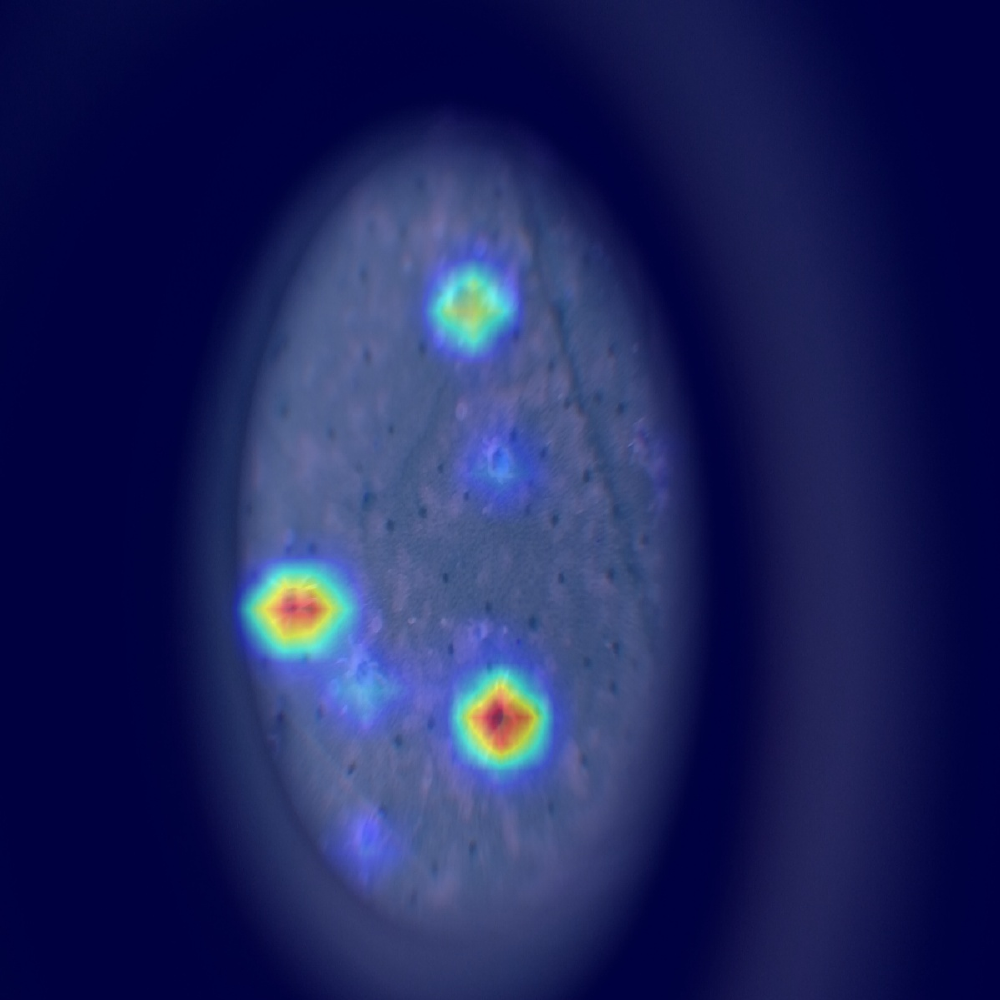}
\hspace*{0.01cm}
\includegraphics[height=1.75cm,width=1.75cm]{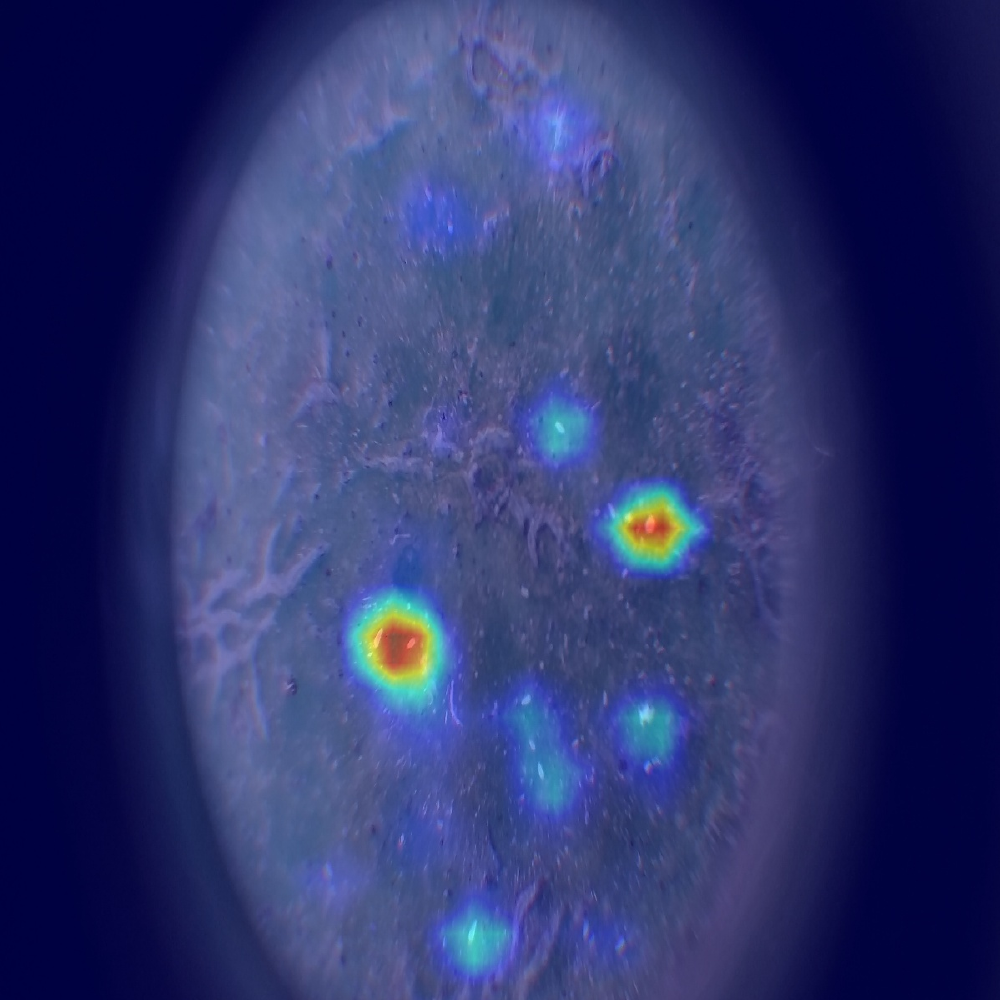}
\hspace*{0.01cm}
\includegraphics[height=1.75cm,width=1.75cm]{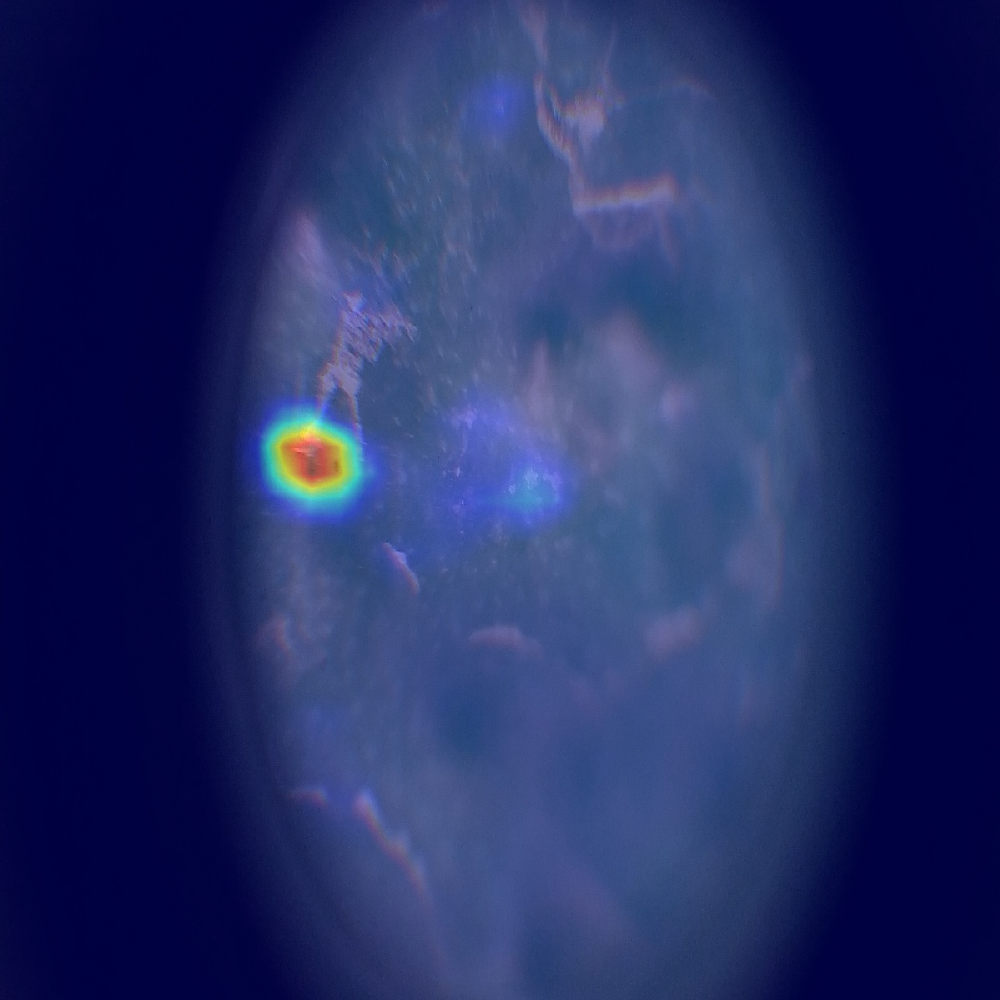}
\hspace*{0.01cm}
\includegraphics[height=1.75cm,width=1.75cm]{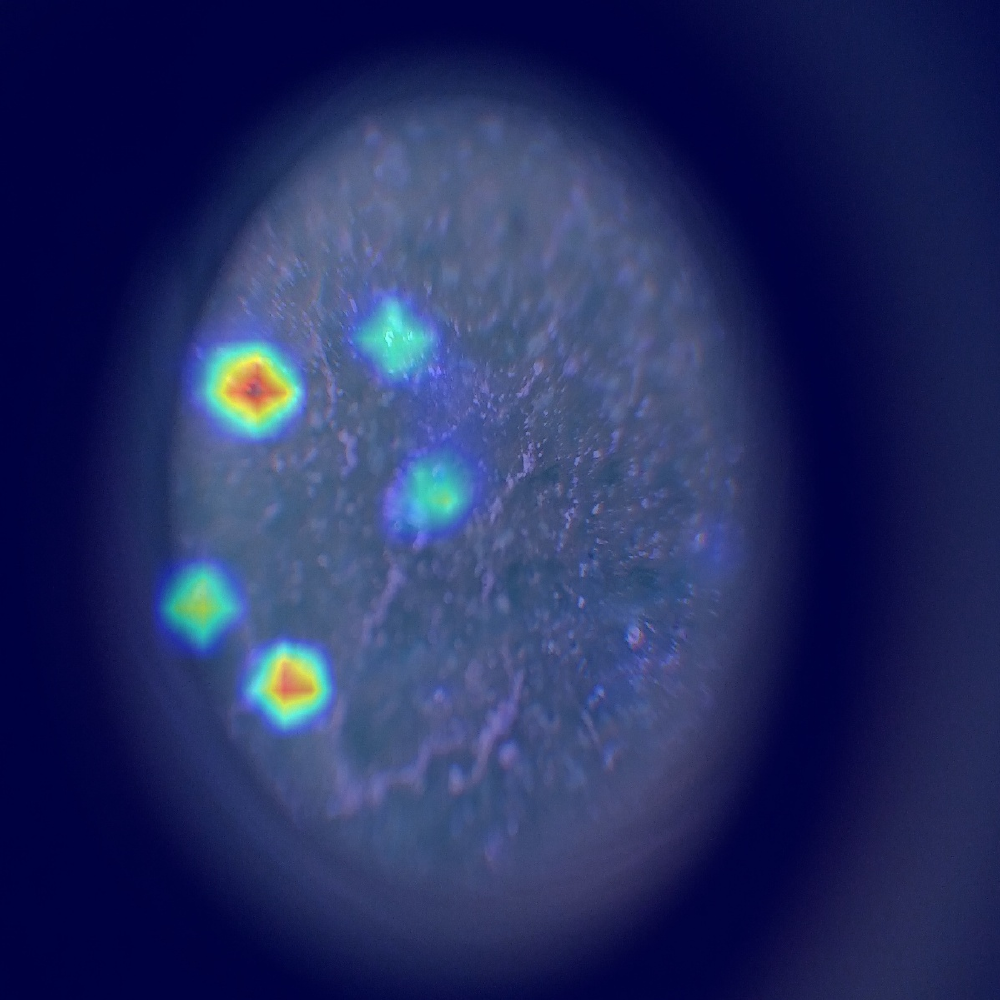}\label{pics:activation-maps-d}} 

\subfloat[Attention-based Deep MIL (instance activations) \citep{ilse2018attention}]{
\includegraphics[height=1.75cm,width=1.75cm]{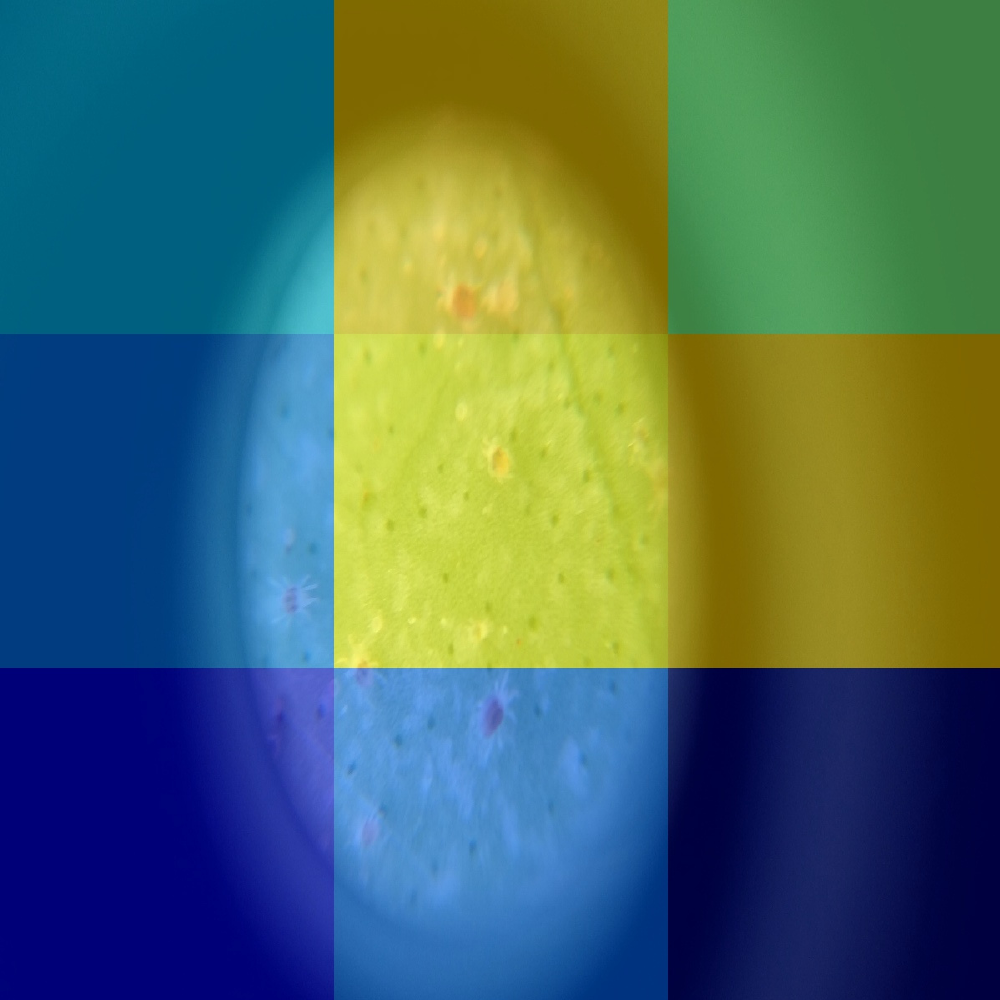} 
\hspace*{0.01cm}
\includegraphics[height=1.75cm,width=1.75cm]{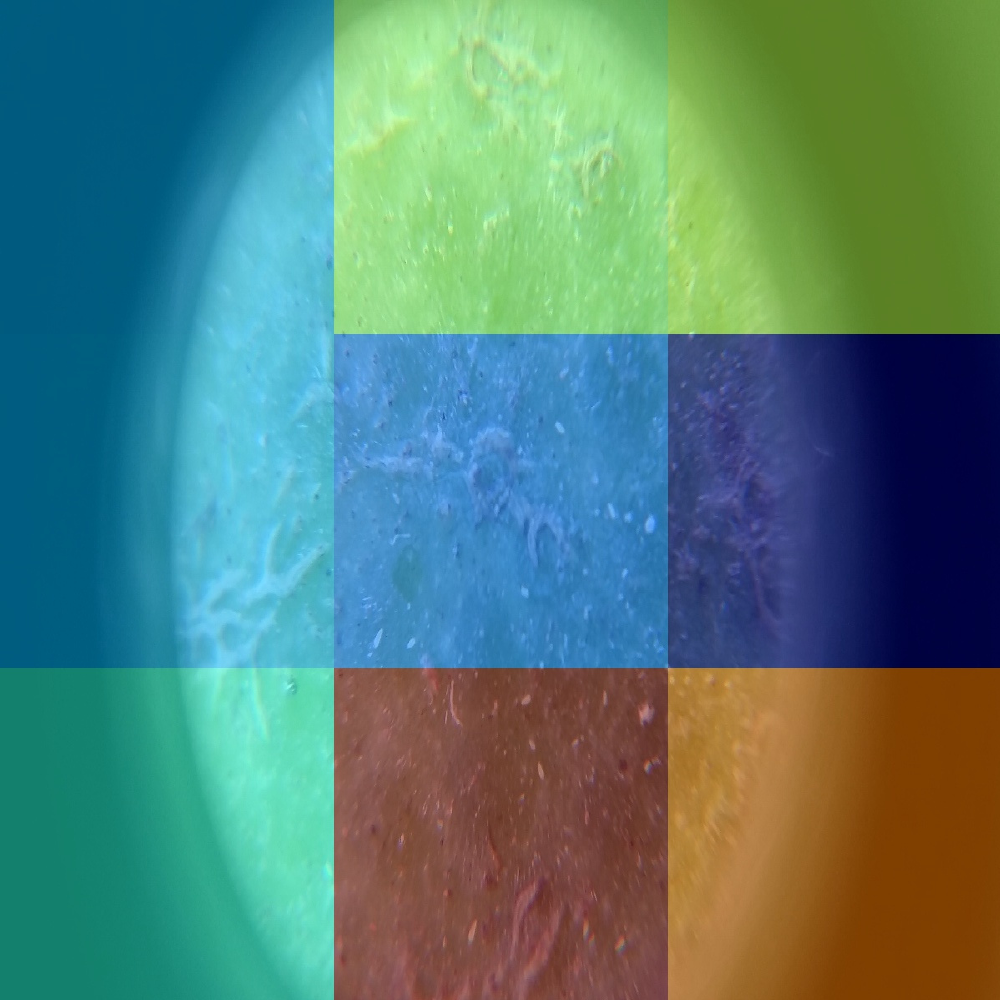}
\hspace*{0.01cm}
\includegraphics[height=1.75cm,width=1.75cm]{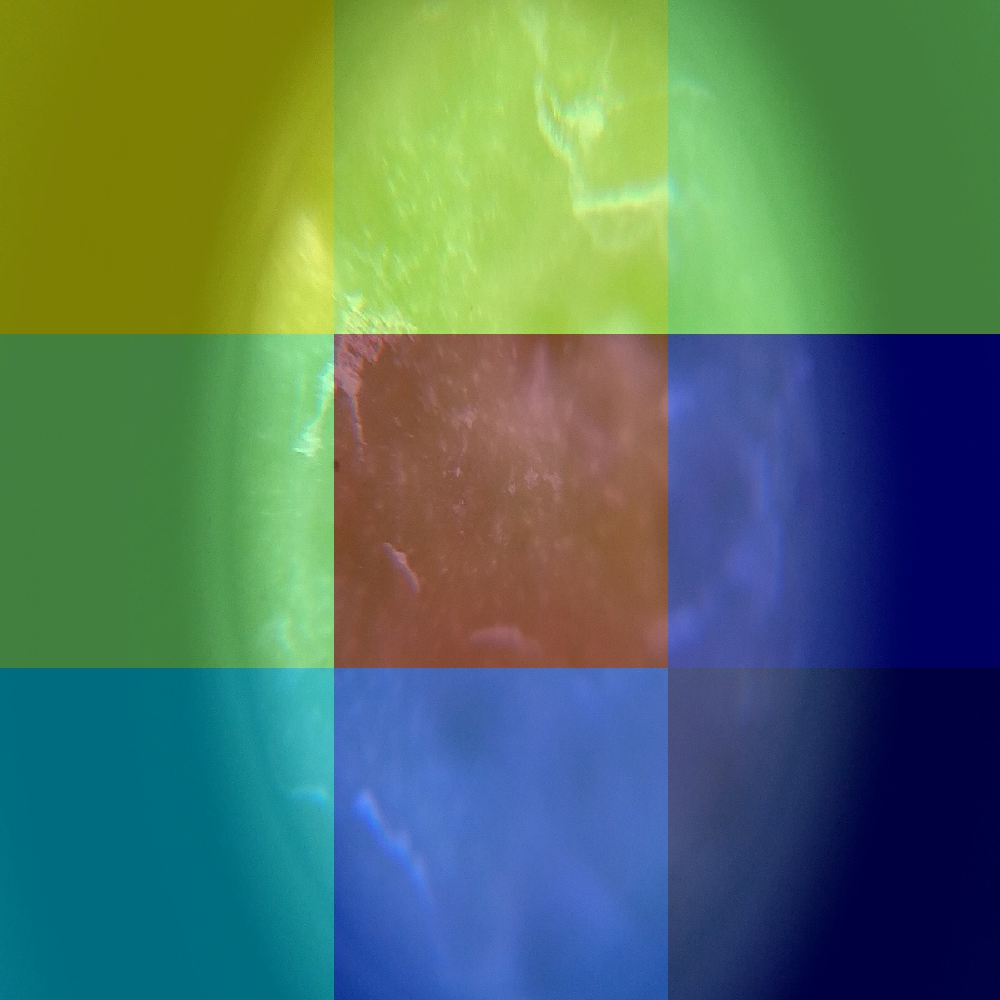}
\hspace*{0.01cm}
\includegraphics[height=1.75cm,width=1.75cm]{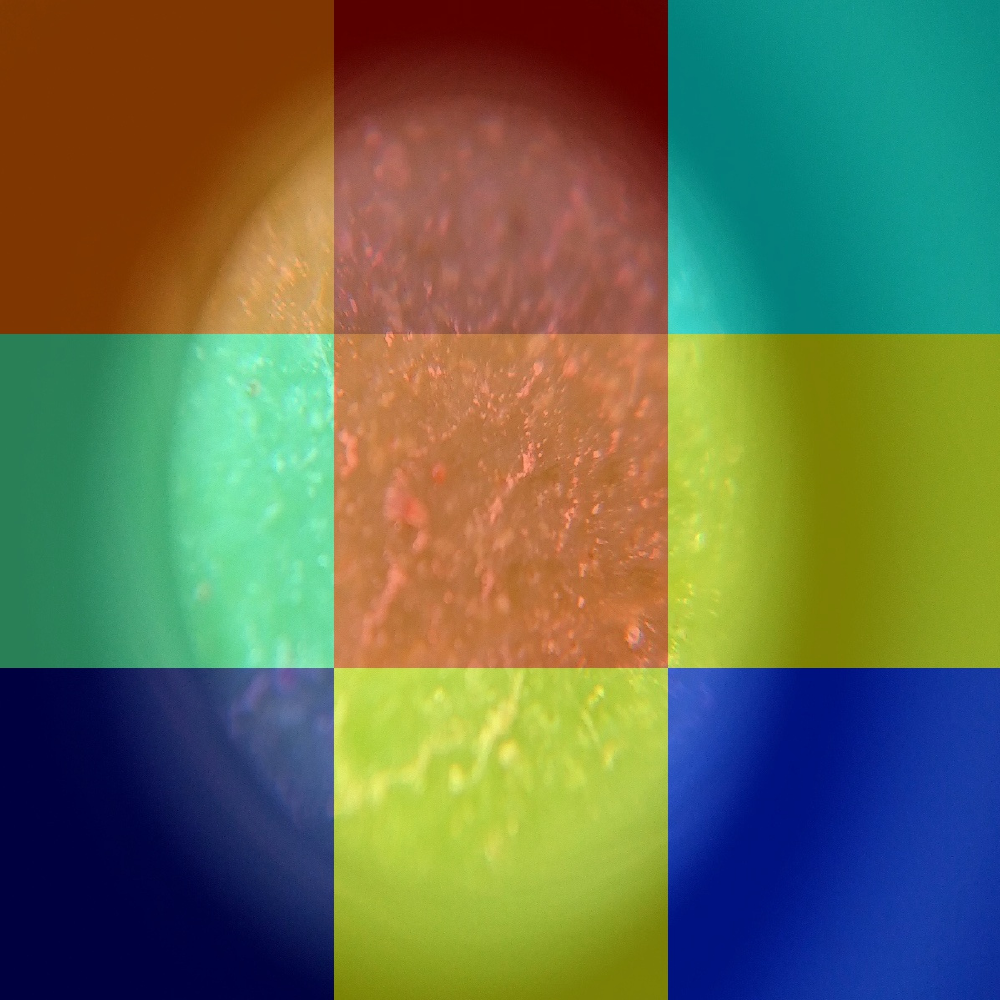}\label{pics:activation-maps-e}}\\

\caption{(a) CPB sample images. (b) Attention-based MIL-Guided saliency maps generated by Two-WAM. (c) MIL-Guided saliency maps generated by Grad-CAM. (d) Saliency maps generated by WILDCAT. (e) Maps based on Attention-based Deep MIL instance activation weights. We generated all activation maps using EfficientNet-B0 as the backbone. Maps from Attention-based Deep MIL show the most important instances for classification tasks, as explained in Subsection~\ref{sec:attention-rel}. The third column illustrates false positive mite locations because it contains a negative class. Red bounding boxes in (a) represent manually annotated mite locations.}
\label{pics:activation-maps}
\end{figure}

In Figure~\ref{pics:activation-maps}, we provide a qualitative comparison for weakly supervised locations produced using EfficientNet-B0 as a backbone. Figure~\ref{pics:activation-maps-a} shows mite locations manually annotated (red bounding boxes), including an image without any mites (third image from left to right).

We present the results of our Attention-based MIL-Guided in Figure~\ref{pics:activation-maps-b}. Two-WAM infers a larger number of salient regions that are more likely to contain mites than the methods in the literature. We notice that it adapts better to the mite bodies, while others overflow the body areas. We believe it is due to the better location of Two-WAM. In Figure~\ref{pics:activation-maps-c}, Grad-CAM red areas (each highlight) often have larger sizes than Attention-based MIL-Guided highlights but, in most cases, the red areas show correct locations. However, the number of mite locations is significantly smaller than the number of bounding boxes. Also, Grad-CAM presents failures in image locations such as the right image of Figure~\ref{pics:activation-maps-c}, where the salient regions are a little bit far from the mite locations. Differently, WILDCAT shows more reliable regions than Grad-CAM with larger areas highlighted in Figure~\ref{pics:activation-maps-d}, but the number of ROIs is still smaller than the number of mites. Figure~\ref{pics:activation-maps-e} shows Attention-based Deep MIL saliency maps created based on each instance attention weight used to classify the entire bag, as explained in Subsection~\ref{sec:attention-rel}. The colors represent how much each instance influenced the final prediction. The patches that contain mites appear in red or yellow color~shades. 

Comparing the regions for the image without mites, we observe that Two-WAM produces more regions, which occurs in the other Two-WAM activation maps. This aspect may be useful for training a robust Instance Model since more samples will be considered. As these samples represent difficult cases for the Bag Model, the Attention-based MIL-Guided method will further analyze the Instance Model to detect these false-positive samples. Although Two-WAM generates more ROIs, their number is not as numerous as the number of mites. Still, Two-WAM is more reliable in identifying the areas where mites are most likely to be located.

\subsection{Weakly Supervised Methods Applied to Insect Pest Dataset}
\label{sec:comparison-IP102}

We evaluate our Attention-based MIL-Guided on Insect Pest dataset, comparing it with MIL-Guided, Attention-based Deep MIL, and WILDCAT methods. IP102 contains images much smaller than CPB, that is, 224$\times$224 pixels. As we explained in Subsection~\ref{sec:IP102}, version~1.1 of IP102 has no published results. Thus, we retrained the literature methods on the new version and present the results in this section. 

We trained all models with the largest batch size possible for each method. For Attention-based Deep MIL, we divide images into 4 instances of 112$\times$112 pixels, 9~instances of 74$\times$74 pixels, and 16 instances of 56$\times$56 pixels. We only reported the best accuracy result since they were not statistically different. Concerning Attention-based MIL-Guided, as the image sizes are smaller than CPB images and ROIs are proportionally larger, we do not evaluate the Instance Model on IP102. For this reason, we present only the Bag Model results.

\begin{figure}[!ht]
\captionsetup[subfloat]{farskip=2pt,captionskip=2pt}
\centering

\subfloat[IP102 samples~\citep{wu2019ip102}]{
\includegraphics[height=1.4cm,width=1.4cm]{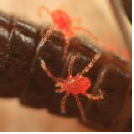} 
\hspace*{0.01cm}
\includegraphics[height=1.4cm,width=1.4cm]{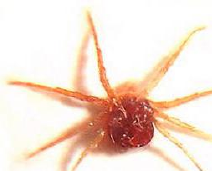}
\hspace*{0.01cm}
\includegraphics[height=1.4cm,width=1.4cm]{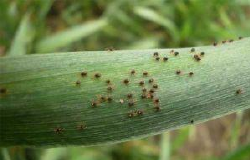}
\hspace*{0.01cm}
\includegraphics[height=1.4cm,width=1.4cm]{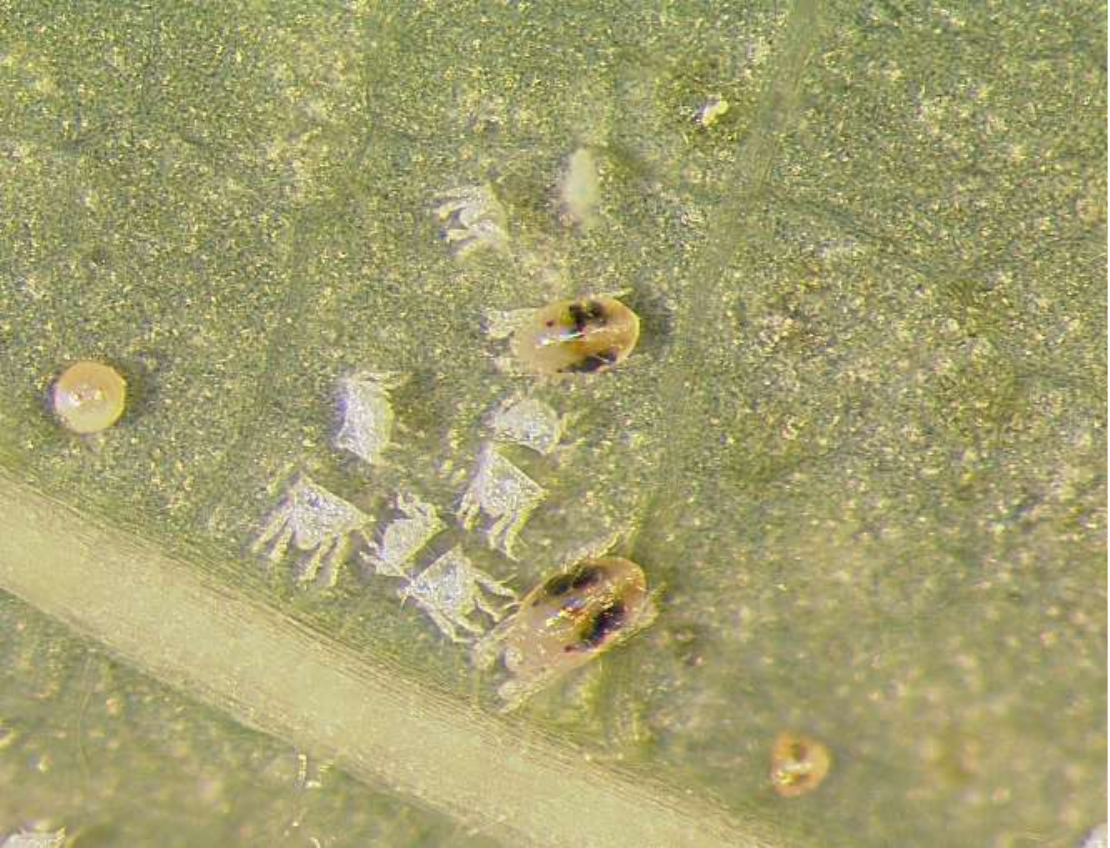}
\hspace*{0.01cm}
\includegraphics[height=1.4cm,width=1.4cm]{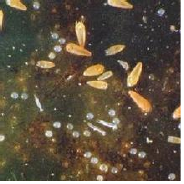}
\label{pics:ip102-activation-maps-a}}\\

\subfloat[Attention-based MIL-Guided (Two-WAM)]{
\includegraphics[height=1.4cm,width=1.4cm]{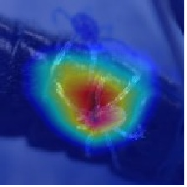}  
\hspace*{0.01cm}
\includegraphics[height=1.4cm,width=1.4cm]{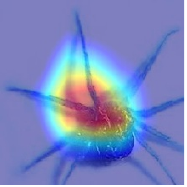}
\hspace*{0.01cm}
\includegraphics[height=1.4cm,width=1.4cm]{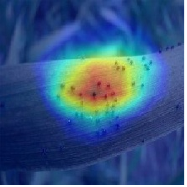}
\hspace*{0.01cm}
\includegraphics[height=1.4cm,width=1.4cm]{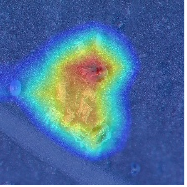}
\hspace*{0.01cm}
\includegraphics[height=1.4cm,width=1.4cm]{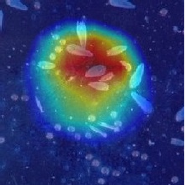}
\label{pics:ip102-activation-maps-b}}  \\

\subfloat[Attention-based MIL-Guided (Two-WAM and CPB Bag Model)]{
\includegraphics[height=1.4cm,width=1.4cm]{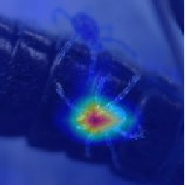}  
\hspace*{0.01cm}
\includegraphics[height=1.4cm,width=1.4cm]{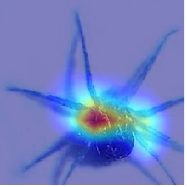}
\hspace*{0.01cm}
\includegraphics[height=1.4cm,width=1.4cm]{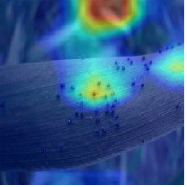}
\hspace*{0.01cm}
\includegraphics[height=1.4cm,width=1.4cm]{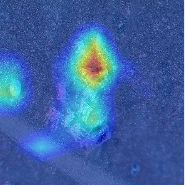}
\hspace*{0.01cm}
\includegraphics[height=1.4cm,width=1.4cm]{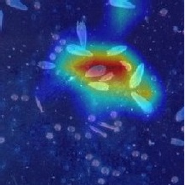}
\label{pics:ip102-activation-maps-c}}  \\

\subfloat[Attention-based MIL-Guided (Two-WAM and CPB Inst. Model)]{
\includegraphics[height=1.4cm,width=1.4cm]{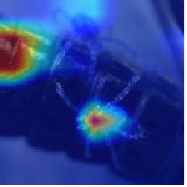}  
\hspace*{0.01cm}
\includegraphics[height=1.4cm,width=1.4cm]{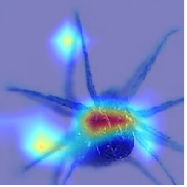}
\hspace*{0.01cm}
\includegraphics[height=1.4cm,width=1.4cm]{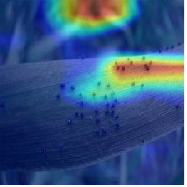}
\hspace*{0.01cm}
\includegraphics[height=1.4cm,width=1.4cm]{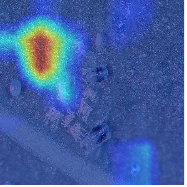}
\hspace*{0.01cm}
\includegraphics[height=1.4cm,width=1.4cm]{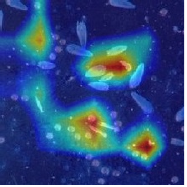}
\label{pics:ip102-activation-maps-d}}

\subfloat[MIL-Guided (Grad-CAM)~\citep{bollis2020weakly}]{
\includegraphics[height=1.4cm,width=1.4cm]{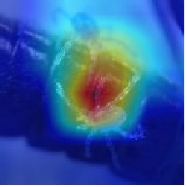}
\hspace*{0.01cm}
\includegraphics[height=1.4cm,width=1.4cm]{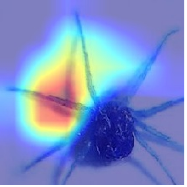} 
\hspace*{0.01cm}
\includegraphics[height=1.4cm,width=1.4cm]{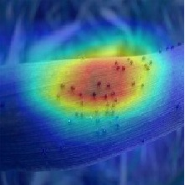}
\hspace*{0.01cm}
\includegraphics[height=1.4cm,width=1.4cm]{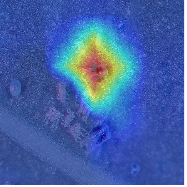}
\hspace*{0.01cm}
\includegraphics[height=1.4cm,width=1.4cm]{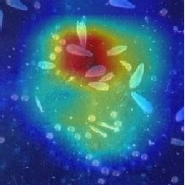}
\label{pics:ip102-activation-maps-e}} \\

\subfloat[WILDCAT~\citep{durand2017wildcat}]{
\includegraphics[height=1.4cm,width=1.4cm]{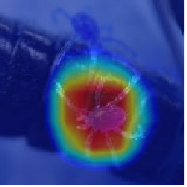}
\hspace*{0.01cm}
\includegraphics[height=1.4cm,width=1.4cm]{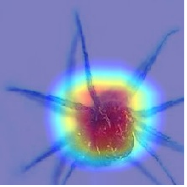}
\hspace*{0.01cm}
\includegraphics[height=1.4cm,width=1.4cm]{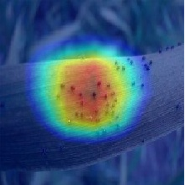}
\hspace*{0.01cm}
\includegraphics[height=1.4cm,width=1.4cm]{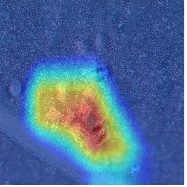}
\hspace*{0.01cm}
\includegraphics[height=1.4cm,width=1.4cm]{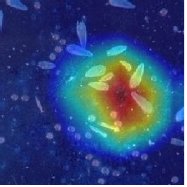}
\label{pics:ip102-activation-maps-f}}\\

\subfloat[Attention-based Deep MIL (instance activations)~\citep{ilse2018attention}]{
\includegraphics[height=1.4cm,width=1.4cm]{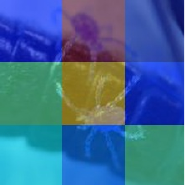} 
\hspace*{0.01cm}
\includegraphics[height=1.4cm,width=1.4cm]{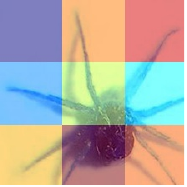}
\hspace*{0.01cm}
\includegraphics[height=1.4cm,width=1.4cm]{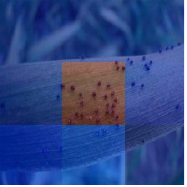}
\hspace*{0.01cm}
\includegraphics[height=1.4cm,width=1.4cm]{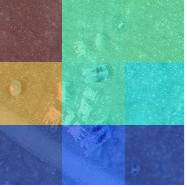}
\hspace*{0.01cm}
\includegraphics[height=1.4cm,width=1.4cm]{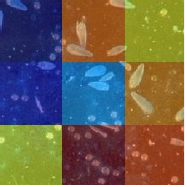}
\label{pics:ip102-activation-maps-g}}

\caption{(a) IP102 sample images \citep{wu2019ip102}. (b) Attention-based MIL-Guided saliency maps generated by Two-WAM. (c) Attention-based MIL-Guided saliency maps generated by Two-WAM using CPB Bag Model weights. (d) Attention-based MIL-Guided saliency maps generated by Two-WAM using CPB Instance Model weights.
(e) MIL-Guided saliency maps generated by Grad-CAM. (f) Saliency maps generated by WILDCAT. (g) Maps based on Attention-based Deep MIL instance activation weights. We generated all activation maps using EfficientNet-B0 as the backbone. Maps from Attention-based Deep MIL show the most important instances for classification tasks, as explained in Subsection~\ref{sec:attention-rel}. In~(c) and~(d), we employed the CPB weights to show that these trained models can infer the mites' locations. Despite some errors, models not trained on CPB identified the IP102 mite locations.}
\label{pics:ip102-activation-maps}
\end{figure}

Table~\ref{table:ReportesIP102} shows that the MIL-Guided achieved the best accuracy and F1-score on IP102, followed by our approach. Attention-based Deep MIL using LeNet achieved the lowest scores, 23.0\% accuracy and 24.5\% F1-score. The results considering 4 and 16 instances using the EfficientNet-B0, not reported in Table~\ref{table:ReportesIP102}, yield 37.8\% accuracy and 37.7\% F1-score, and 36.3\% accuracy and 36.5\% F1-score on average, respectively. One possible explanation for such a result of Attention-based Deep MIL is that the ROIs on IP102 are often larger than on CPB images. Each instance contains only part of these large regions, as illustrated in Figure~\ref{pics:ip102-activation-maps-g}.  Likewise, Instance Models present worse results on IP102.

\begin{table}[t]
\setlength{\tabcolsep}{1.0mm}
\begin{center}
\caption{Classification accuracy (Acc. in \%) and F1-score (F1. in \%) of different WSL on IP102 test set. The highlights in bold correspond to the best results.}
\vspace{0.1cm}
\label{table:ReportesIP102}
\footnotesize
\begin{tabular}{lcc}
\toprule
                    & \multicolumn{2}{c}{$224\times224$}  \\
WSL &  Acc. (\%) & F1 (\%) \\
\midrule
Attention-based Deep MIL (LeNet)~\citep{ilse2018attention}  & $23.0$ \tiny{$\pm 1.1$}  & $24.5$ \tiny{$\pm1.0$} \\
Attention-based Deep MIL (EfficientNet) & $38.7$ \tiny{$\pm 3.3$}  & $36.8$ \tiny{$\pm 3.8$}\\
WILDCAT (ResNet101)~\citep{durand2017wildcat} & $67.6$ \tiny{$\pm0.4$} & $67.5$	\tiny{$\pm0.6$}\\
WILDCAT (EfficientNet)  & $65.1$ \tiny{$\pm 0.4$}  & $64.4$ \tiny{$\pm 0.9$}\\
MIL-Guided (Bag Model)~\citep{bollis2020weakly}  & $\textbf{69.5}$ \tiny{$\pm \textbf{0.1}$} & $\textbf{69.0}$ \tiny{$\pm \textbf{0.1}$} \\
Attention-based MIL-Guided (Bag Model) & $68.3$ \tiny{$\pm 0.3$} & $68.0$ \tiny{$\pm 0.1$}\\

\bottomrule
\end{tabular}
\end{center}
\end{table}

Regarding WILDCAT, which uses entire images in training, their results are closer to MIL-Guided and Attention-based MIL-Guided. It means 65.1\% accuracy and 64.4\% F1-score with EfficientNet-B0, and 67.6\% accuracy and 67.5\% F1-score with ResNet101. MIL-Guided (our previous work) surpassed our Attention-based MIL-Guided by 1 percentage point. Both Bag Models outperformed both methods from literature by at least 2 percentage points.

\begin{figure*}[t]
\captionsetup[subfloat]{farskip=2pt,captionskip=2pt}
\centering

\subfloat[IP102 samples~\citep{wu2019ip102}]{
\includegraphics[height=1.5cm,width=1.5cm]{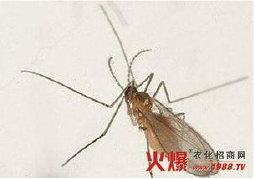} 
\includegraphics[height=1.5cm,width=1.5cm]{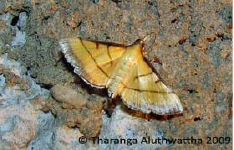}
\includegraphics[height=1.5cm,width=1.5cm]{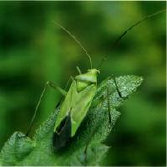}
\includegraphics[height=1.5cm,width=1.5cm]{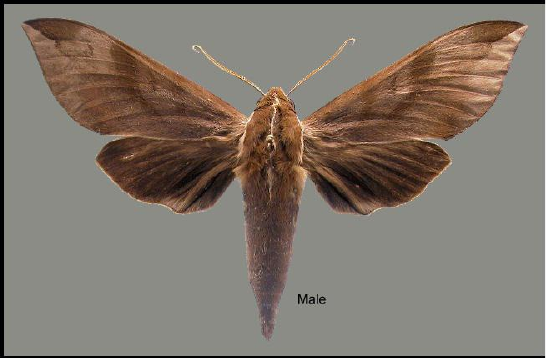}
\includegraphics[height=1.5cm,width=1.5cm]{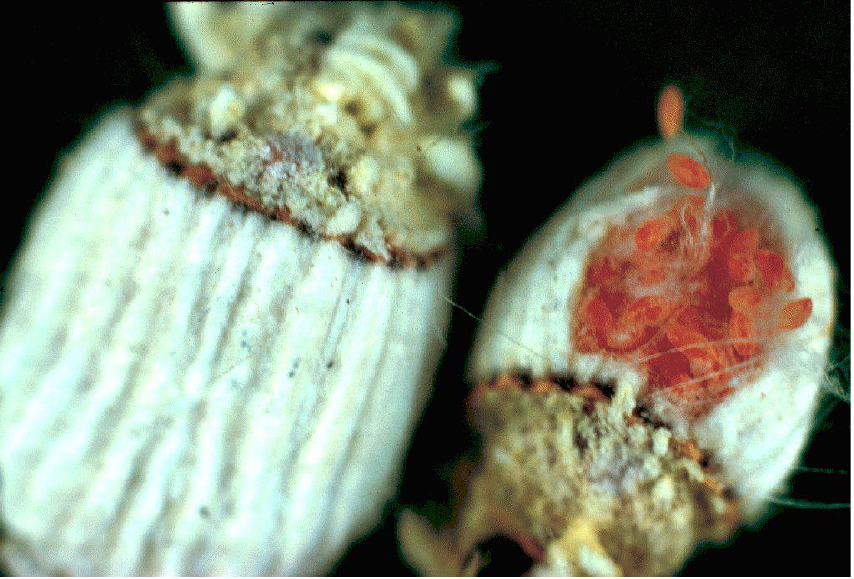}
\includegraphics[height=1.5cm,width=1.5cm]{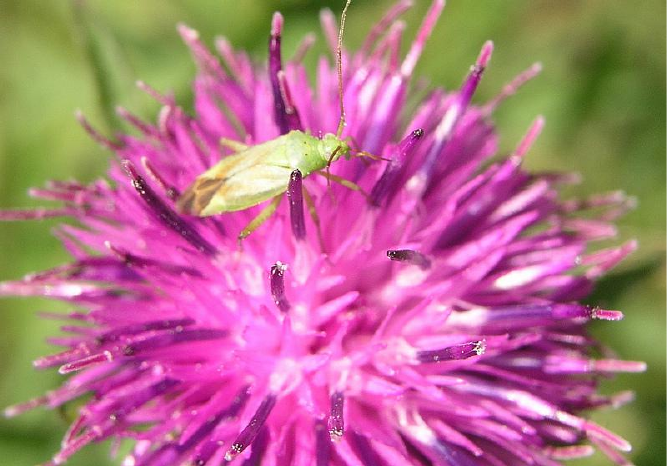}
\includegraphics[height=1.5cm,width=1.5cm]{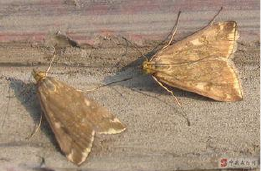}
\includegraphics[height=1.5cm,width=1.5cm]{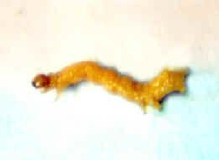}
\includegraphics[height=1.5cm,width=1.5cm]{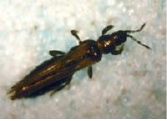}
\includegraphics[height=1.5cm,width=1.5cm]{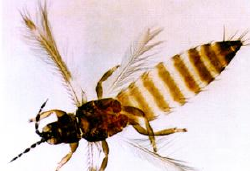}
\label{pics:ip-102-a}}\\

\subfloat[Attention-based MIL-Guided (Two-WAM)]{
\includegraphics[height=1.5cm,width=1.5cm]{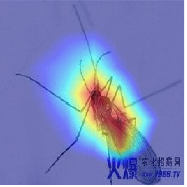} 
\includegraphics[height=1.5cm,width=1.5cm]{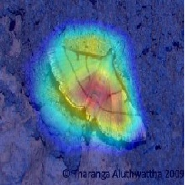}
\includegraphics[height=1.5cm,width=1.5cm]{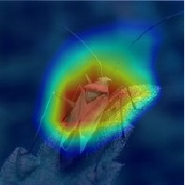}
\includegraphics[height=1.5cm,width=1.5cm]{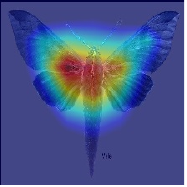}
\includegraphics[height=1.5cm,width=1.5cm]{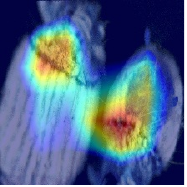}
\includegraphics[height=1.5cm,width=1.5cm]{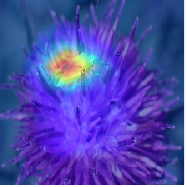}
\includegraphics[height=1.5cm,width=1.5cm]{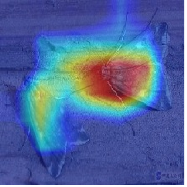}
\includegraphics[height=1.5cm,width=1.5cm]{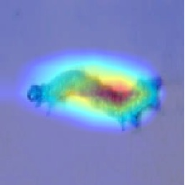}
\includegraphics[height=1.5cm,width=1.5cm]{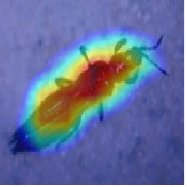}
\includegraphics[height=1.5cm,width=1.5cm]{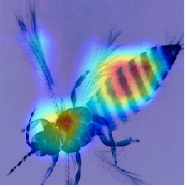}

\label{pics:ip-102-b}} \\

\subfloat[MIL-Guided (Grad-CAM)~\citep{bollis2020weakly}]{
\includegraphics[height=1.5cm,width=1.5cm]{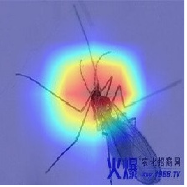} 
\includegraphics[height=1.5cm,width=1.5cm]{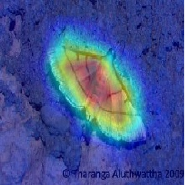}
\includegraphics[height=1.5cm,width=1.5cm]{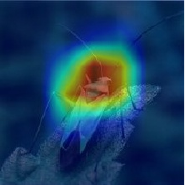}
\includegraphics[height=1.5cm,width=1.5cm]{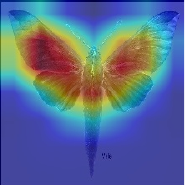}
\includegraphics[height=1.5cm,width=1.5cm]{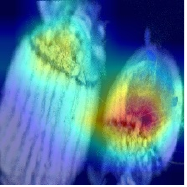}
\includegraphics[height=1.5cm,width=1.5cm]{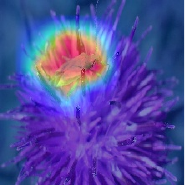}
\includegraphics[height=1.5cm,width=1.5cm]{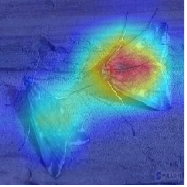}
\includegraphics[height=1.5cm,width=1.5cm]{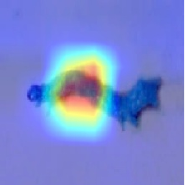}
\includegraphics[height=1.5cm,width=1.5cm]{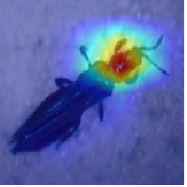}
\includegraphics[height=1.5cm,width=1.5cm]{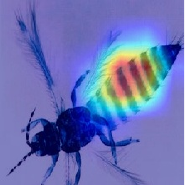}
\label{pics:ip-102-c}}\\

\subfloat[Attention-based MIL-Guided (Two-WAM) bounding boxes]{
\includegraphics[height=1.5cm,width=1.5cm]{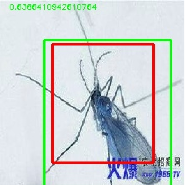} 
\includegraphics[height=1.5cm,width=1.5cm]{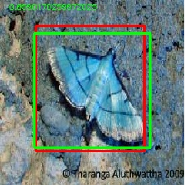}
\includegraphics[height=1.5cm,width=1.5cm]{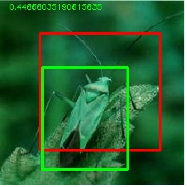}
\includegraphics[height=1.5cm,width=1.5cm]{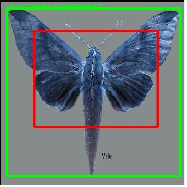}
\includegraphics[height=1.5cm,width=1.5cm]{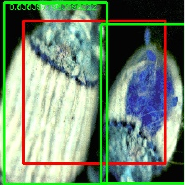}
\includegraphics[height=1.5cm,width=1.5cm]{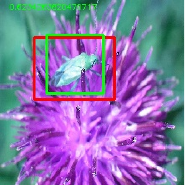}
\includegraphics[height=1.5cm,width=1.5cm]{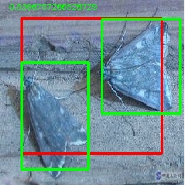}
\includegraphics[height=1.5cm,width=1.5cm]{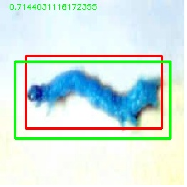}
\includegraphics[height=1.5cm,width=1.5cm]{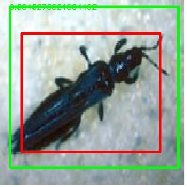}
\includegraphics[height=1.5cm,width=1.5cm]{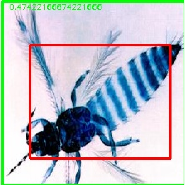}

\label{pics:ip-102-d}}   \\

\subfloat[MIL-Guided (Grad-CAM) bounding boxes~\citep{bollis2020weakly}]{
\includegraphics[height=1.5cm,width=1.5cm]{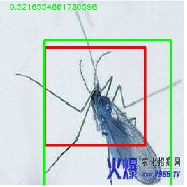} 
\includegraphics[height=1.5cm,width=1.5cm]{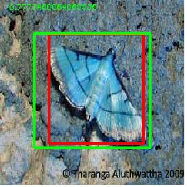}
\includegraphics[height=1.5cm,width=1.5cm]{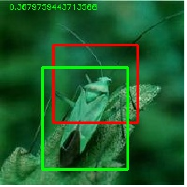}
\includegraphics[height=1.5cm,width=1.5cm]{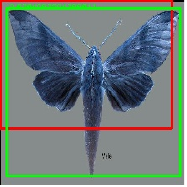}
\includegraphics[height=1.5cm,width=1.5cm]{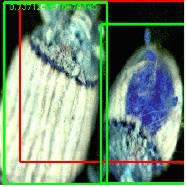}
\includegraphics[height=1.5cm,width=1.5cm]{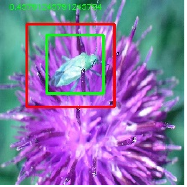}
\includegraphics[height=1.5cm,width=1.5cm]{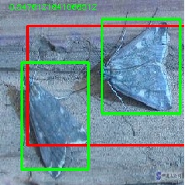}
\includegraphics[height=1.5cm,width=1.5cm]{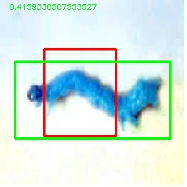}
\includegraphics[height=1.5cm,width=1.5cm]{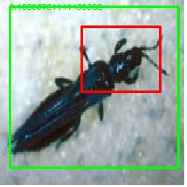}
\includegraphics[height=1.5cm,width=1.5cm]{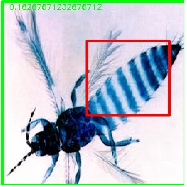}

\label{pics:ip-102-e}} 

\caption{(a) IP102 sample images. (b) MIL-Guided saliency maps produced by Grad-CAM. (c) Attention-based MIL-Guided saliency maps based on Two-WAM. (d) MIL-Guided bounding boxes produced. (e) Attention-based MIL-Guided bounding boxes. Green boxes are the ground truth and red are inferences. Values in green are intersection over unions (IoU).}
\label{pics:ip-102}
\end{figure*}

Figure~\ref{pics:ip-102} illustrates why Two-WAM reduces the classification performance on larger insect images. As the literature describes~\citep{selvaraju2017grad} and Figure~\ref{pics:ip-102-c} shows, Grad-CAM saliency maps evidence areas that strongly influence final predictions. This means that Grad-CAM often highlights areas with a small number of regions (red areas) where the most descriptive features for classification are located. On the other hand, Figure~\ref{pics:ip-102-b} (Two-WAM) highlights different areas, but they have an intersection with Grad-CAM ones. Therefore, the Attention-based MIL-Guided regions may also contain features that are not discriminative for classification.

In addition, the Grad-CAM highlights on the red areas are centered on some insect regions and commonly exceed the insect’s bodies. In contrast, the regions covered by Two-WAM are drawn along the insect body areas and fit better to those bodies. Grad-CAM greater red areas may indicate that borders are essential for producing classification scores. Therefore, as mentioned before, our results suggest that the localization capacity to fill the entire body of the insect or mite does not constantly improve the classification. Two-WAM showed more than one red area in some cases, whereas Grad-CAM showed only connected regions for two insects. Consequently, Two-WAM improved the weakly supervised localization but decreased accuracy by 1 percentage point over MIL-Guided.

Figure~\ref{pics:ip-102-d} shows the Two-WAM localization performance in generating bounding boxes based on its activation maps. Figure~\ref{pics:ip-102-e}, which exhibits Grad-CAM bounding boxes, shows less accurate areas, sometimes containing only a small part of insect bodies. Resuming the Two-WAM capacity, it generates more precise bounding boxes and infers many regions to provide more reliable results, all without training with location labels. However, our bounding box generator, inspired by~\citet{Lu2017a}, does not yield more than one bounding box per image.

Figure~\ref{pics:ip102-activation-maps} exemplifies saliency maps from IP102 mites. It is worth pointing out that some images have smaller ROIs, but they are not as small as CPB mite images. Attention-based MIL-Guide using Two-WAM in Figure~\ref{pics:ip102-activation-maps-b} often highlights all mites, but similarly to the other methods, except for Attention-based Deep MIL (Figure~\ref{pics:ip102-activation-maps-g}), it does not localize one of the mites in the first column, and it does not highlight the entire mite in the second column. MIL-Guided using Grad-CAM (Figure~\ref{pics:ip102-activation-maps-e}) focuses on one region and it does not highlight entire areas as usual. The WILDCAT saliency maps (Figure~\ref{pics:ip102-activation-maps-f}) show more concise regions and fit better to the mite groups. However, in cases where mites are scattered, the marked regions do not cover them entirely and overflow the mite borders. Figure~\ref{pics:ip102-activation-maps-g} illustrates more activated instances that influenced Attention-based Deep MIL predictions, but never highlights all instances. 

Moreover, as illustrated in Figures~\ref{pics:ip102-activation-maps-c} and~\ref{pics:ip102-activation-maps-d}, we investigated whether our CPB models correctly infer ROIs in mites from IP102. This comparison shows that our CPB models can identify the correct location of mites in a dataset generated with requirements different from the CPB requirements. However, training in another base made possible the appearance of errors such as those in the middle images in Figures~\ref{pics:ip102-activation-maps-c}~and~\ref{pics:ip102-activation-maps-d}. This means our CPB models learned when mites are present on the image and where they are, but only highlighted small regions because they were trained in a dataset with small regions. In the last column in  Figure~\ref{pics:ip102-activation-maps-d}, the Instance Model trained on CPB instances evidenced each small group of mites, maintaining a finer-grained demarcation than the other methods. The mites shown in the last column in Figure~\ref{pics:ip102-activation-maps} are rust mites (\textit{Phyllocoptruta oleivora}), also present in the CPB.   

\subsection{Discussions}
\label{sec:discussion}

We performed several experiments to determine whether attention-based activation maps effectively classify pests under images taken from natural conditions. In some scenarios, field images or noise images without any pre-processing were essential for the classification task. Experiments based on these scenarios handle salient ROIs, as most approaches in the literature~\citep{wu2019ip102, Wang2020a}. However, our experiments using tiny ROIs showed that noise in the training process results in less effective models.

The proposed Two-WAM generated more accurate weakly supervised locations than Grad-CAM. However, it reduced the classification performance by 1 percentage point in salient regions. This behavior can be explained by analyzing the saliency maps from Grad-CAM on the IP102 dataset, which shows only specific areas in red responsible for the predictions in most sampled images. On the other hand, Two-WAM achieved more reliable insect locations, but the features produced are insufficient to improve classification performance. On CPB, as ROIs are tiny, Two-WAM generated a more significant number of correct mite regions. Consequently, it refined instance generation and provided better patches to the Instance Model, which improved the performance rates by at least 3 percentage points in the 800$\times$800 scenario and 1 in the 1200$\times$1200 scenario.

Concerning Attention-based MIL-Guided as a process, we conducted experiments to train the Bag Model and the Instance Model in an end-to-end pipeline. The results are not as good as the Bag Model's predictions so far. We understood that the step-by-step process was responsible for improving results in tiny ROIs.

Furthermore, we tried to increase the feature map areas at the end of the convolutions, but it did not change the model's classification accuracy. The results suggest that increasing the feature map sizes does not influence or modify the Two-WAM locations. Finally, we tested different values for the parameter $c$ in Equation~\ref{eq:trans}, but this did not provide significant improvements.

\section{Conclusions and Future Work}
\label{sec:conclusions}

In this work, we proposed a new attention-based activation map approach, called Two-Weighted Activation Map (Two-WAM), to improve a weakly supervised learning process known as Multiple Instance Learning Guided by Saliency Maps  (MIL-Guided)~\citep{bollis2020weakly}. We introduced the Attention-based Multiple Instance Learning Guided by Saliency Maps (Attention-based MIL-Guided). We analyzed the training influence of noisy images with small and salient regions of interest on deep neural networks.

We conducted the experiments on two challenging datasets, Citrus Pest Benchmark (CPB)~\citep{bollis2020weakly} and Insect Pest dataset (IP102)~\citep{wu2019ip102}. We compared the Attention-based MIL-Guided with MIL-Guided~\citep{bollis2020weakly} (our previous work) and two state-of-the-art methods, Attention-based Deep MIL~\citep{ilse2018attention} and WILDCAT~\citep{durand2017wildcat}. 
We~qualitatively compared activation maps produced by Two-WAM (Attention-based MIL-Guided), Grad-CAM (MIL-Guided), Attention-based Deep MIL, and WILDCAT.

Our results showed that the noise in tiny images disturbed the training process and negatively influenced the models' classification performance. On the other hand, noise in salient images (i.e., in the Instance Model) helped training the models. We reached the best mark on the test set concerning the Attention-based MIL-Guided results against MIL-Guided, Attention-based Deep MIL, and WILDCAT trained on CPB. We improved state of the art achieving 92.4\% accuracy and 91.8\% F1-score, and our results surpassed Attention-based Deep MIL and WILDCAT in all scenarios by up to 25.3 percentage points.  

Two-WAM showed a better ability to highlight mite areas than literature methods (Attention-based Deep MIL, Grad-CAM, and WILDCAT), increasing the correct amount of mites pointed to generate instances. Regarding IP102 insects and mites, Two-WAM created more reliable activation maps, with red areas better adapted to pest bodies and highlighted a greater amount of these areas. However, the consequence of improving the maps was the slight drop in the classification performance of 1 percentage point with our best result concerning MIL-Guided reaching 69.5\% accuracy and 69.0\% F1-score. Attention-based MIL-Guided and MIL-Guided achieved superior results than Attention-based Deep MIL and WILDCAT.
 
As for directions for future work, we plan to make Two-WAM class-dependent by calculating different weights for each category. Our next step is to apply the Attention-based MIL-Guided process in a multi-class task to classify mite species. We will investigate the use of the Attention-based MIL-Guided to automate the integrated pests management (IPM) process using mobile phones and magnifying glasses. As far as we know, there are no examples of non-toy apps being used to recognize mites in orchards due to the inherent challenges of automatic classification.

Attention-based MIL-Guided shows better results in classifying tiny regions than other WSL methods in the literature, even with few parameters. We plan to deploy our models on mobile devices and use them in the field. It is noteworthy that new pest benchmarks have addressed tiny regions~\citep{pei2020enhancing, Wang2021, lins2020method}. Introducing images with small ROIs may become frequent in pest datasets because pests often represent small areas in uncontrolled field images.

\section{Credit Authorship Contribution Statement}

EB performed all experiments. EB, HM, HP, and~SA performed data analysis. EB, HM, HP, and SA wrote~the manu\-script. EB, HM, HP, and SA performed~the manuscript proofreading and prepared tables and figures.

\section{Acknowledgements}

EB is partially funded by CAPES (88882.329130/2019-01). HM, HP, SA are partially funded by H.IAAC (Artificial Intelligence and Cognitive Architectures Hub). HP is also partially funded by FAPESP (2017/12646-3) and CNPq (309330/2018-1). SA is also partially funded by CNPq PQ-2 (315231/2020-3), FAPESP (2013/08293-7), and Google Research Awards for Latin America 2020. RECOD Lab. is partially supported by diverse projects and grants from FAPESP, CNPq, and CAPES. We gratefully acknowledge the donation of GPUs by NVIDIA Corporation. 

\section{Declaration of Competing Interest}

The authors declare that they have no known competing financial interests or personal relationships that could have appeared to influence the work reported in this paper.

\bibliography{references}



\end{document}